\documentclass[11pt]{article}

\usepackage[margin=1in]{geometry}
\usepackage{amsmath,amssymb}
\usepackage{booktabs}
\usepackage{array}
\usepackage{graphicx}
\usepackage{xurl}
\usepackage{hyperref}
\usepackage{xcolor}
\usepackage{enumitem}
\usepackage{microtype}
\usepackage{float}
\usepackage{placeins}
\usepackage{pdflscape}
\graphicspath{{figures/}}

\setlist{nosep,leftmargin=1.5em}
\hypersetup{
  pdftitle={Logical Judgments Under Pressure: Diagnosing Syllogistic Stability with Learned Soft Prefixes},
  pdfauthor={Brian K Chen},
  colorlinks=true,
  linkcolor=blue,
  citecolor=blue,
  urlcolor=blue
}

\title{Logical Judgments Under Pressure: Diagnosing Syllogistic Stability with Learned Soft Prefixes}
\author{Brian K Chen\\National University of Singapore\\\texttt{e0694208@u.nus.edu}}
\date{}

\newcolumntype{P}[1]{>{\raggedright\arraybackslash}p{#1}}

\begin{document}
\maketitle

\begin{abstract}
To test how correct logical judgments respond to learned context, we prepend a soft prefix to an exactly labeled syllogistic reasoning benchmark while keeping the model fixed. Soft prefixes are opaque continuous vectors, so we characterize them through the behavior they induce across controlled variations in logical form and interface. By studying which prefixes succeed and how their effects generalize, we characterize how learned contextual pressure can override correct judgments and expose limits in a model's logical stability. Across Qwen3.6-35B-A3B MoE, Qwen3-8B, and Gemma 4 31B, learned prefixes redirect many correct answers and remain effective across unseen forms and interface changes. In repeated tests with Qwen3.6 MoE and Gemma, they outperform paired random controls in all 16 model--direction--split comparisons by 37 to 99 percentage points. Qwen3.6 MoE flip rates remain between 72\% and 90\% across wording and prompt changes, while Gemma validity prefixes retain 54\% to 56\% flip compared with less than 1\% for matched random prefixes. Diagnostic tests show that the dominant effect is a broad preference for one answer meaning rather than fixed-symbol forcing or a logical operation that transfers reliably between tasks. The form of this bias differs across models. In both Qwen models, simple score models often predict which judgments will flip but not how far their margins will move, whereas Gemma's overall response is more closely approximated by the same models. These results show that the dominant behavioral effect of successful soft prefixes is a broad answer preference, while the remaining response reveals substantial model-specific differences in logical stability.

\end{abstract}

\section{Introduction}

Reliable logical reasoning requires more than producing the correct answer under a single prompt. A model may make the correct formal judgment yet reverse it when context outside the problem changes, even though the logical content is fixed. We therefore distinguish accuracy on logical tasks from robustness. Accuracy asks whether a judgment is correct, whereas robustness asks whether it remains stable under added context. Benchmark accuracy alone does not reveal what kinds of learned context can overcome otherwise correct logical behavior.

We study this question using syllogistic logic, where labels can be computed exactly. We prepend a short trainable sequence of embedding vectors to every prompt while keeping the language model fixed. The prefix is optimized to redirect a chosen class of judgments that the unprefixed model answers correctly toward a specified target answer. Because soft prefixes are continuous vectors rather than readable instructions, their meaning cannot be inspected directly. We instead interpret them functionally by asking which judgments they change, where their effects transfer, and which simpler mechanisms reproduce them. By identifying what effective prefixes do, we can infer what kinds of learned context can override correct logical judgments and where a model's stability boundaries lie.

We focus on unsatisfiable-to-satisfiable and valid-to-invalid reversals for a logical reason. Even if the prefix were interpreted as additional premises, neither reversal is licensed under the benchmark semantics. Adding information cannot make an inconsistent set satisfiable, and monotonic entailment cannot become invalid merely because more information is added. A successful prefix in these directions therefore redirects the model's judgment rather than supplying a legitimate premise, proof, or counterexample.

These directions create an important interpretive ambiguity. Both move a minority answer toward the majority answer, and every example from the other class already has the target label. An always-target system would therefore flip every targeted answer while preserving the other class. High flip with low damage can establish that an intervention is strong and transfers to unseen logical forms, but it cannot establish selective logical editing. The central question is therefore what kind of behavior produces this pattern.

We distinguish three possibilities. A prefix may force a fixed output symbol, create a broad preference for the target answer within one task, or implement a more general operation that transfers between validity and satisfiability. The unchanged formal problem and exact semantics keep the correct answer fixed throughout these tests. Randomized answer mappings test for fixed-symbol forcing. Target-answer rates, direct score bonuses, and fitted score models test for broad answer preference and measure how consistently it acts across examples. Cross-task transfer tests whether the prefix implements a shared operation across the two tasks. Splits by logical form, new wordings, and rephrased prompts test how well the effect generalizes. Random-prefix and readable-text controls test whether it reflects generic sensitivity to added context, while reverse directions reveal its directional limits. Supporting activation analyses then examine where the final-answer change can be interrupted. Together, these experiments give otherwise opaque prefixes a functional interpretation without treating them as readable instructions or logical edits.

Across Qwen3.6-35B-A3B MoE~\cite{qwen36}, Qwen3-8B~\cite{qwen3}, and Gemma 4 31B~\cite{gemma4}, learned prefixes redirect many previously correct judgments on unseen logical forms and remain effective when wording or prompt structure changes. Across the new-wording and rephrased-prompt conditions, flip rates range from 72\% to 90\% for Qwen3.6 MoE, 53\% to 72\% for Qwen3-8B, and 40\% to 56\% for Gemma 4 31B. Across four different partitions of the logical forms, learned prefixes outperform paired random controls in all 16 Qwen3.6 and Gemma comparisons, with advantages of 37 to 99 percentage points. The matched transfer results are especially clear for Gemma validity, where learned prefixes retain 54\% to 56\% flip under new wordings and rephrased prompts, while matched random prefixes remain below 1\%. A complementary Qwen3.6 test shows that even the strongest prefixes selected from 1{,}000 random candidates transfer substantially less than the learned prefixes.

The diagnostic profile shows that the dominant effect is a broad preference for one answer meaning rather than a fixed output symbol or a logical operation that transfers reliably between the two tested tasks. This broad answer preference explains much of the aggregate effect, but it does not fully explain how strongly individual examples respond. Score models that use answer class and the unprefixed margin approximate Gemma's overall prefixed margins substantially better than Qwen's. For both Qwen models, despite their different scale and architecture, these models often predict which judgments will cross the decision boundary but generally do not predict their final margins. Similar aggregate steering rates can therefore conceal different response patterns across models.

Our contributions are:
\begin{itemize}
    \item We introduce a controlled soft-prefix stress test for exactly labeled syllogistic judgments, using logically motivated intervention directions, splits by logical form, and matched random controls.
    
    \item We give opaque continuous prefixes a functional interpretation by distinguishing fixed-symbol forcing, broad semantic answer bias, transfer across forms and interfaces, and directional or task-specific limits.
    
    \item We identify a cross-model difference that aggregate flip rates conceal. Score models based on answer class and the unprefixed margin approximate Gemma's overall response better than Qwen's, while often predicting Qwen's binary flips without predicting its final targeted margins. Supporting activation analyses show that learned prefixes produce much larger answer-margin changes than norm-matched random prefixes, while sensitivity to prefix order differs across models.
\end{itemize}

\section{Related Work}

\paragraph{Syllogistic reasoning and reasoning biases in language models.}
Syllogisms are useful probes because they combine exact logical labels with well-studied reasoning biases. Prior work shows that language-model judgments vary with semantic content, known biases, prompt design, and fine-tuning~\cite{dasgupta2022content,ando2023neubaroco,eisape2023systematic,bertolazzi2024softreasoners}. Biomedical and multilingual benchmarks likewise show the importance of vocabulary and prompt design~\cite{wysocka2024syllobio,nguyen2025bis}. We ask a complementary question: when a model is already correct, how does its answer respond to optimized context outside the formal problem?

\paragraph{Accuracy and answer robustness.}
Our work is closest in motivation to Dong, Jamnik, and Li{\`o}, who show that high syllogistic accuracy can coexist with incorrect explanations and sensitivity to surface form~\cite{dong2026datadriven}. Strong performance on controlled fragments of language can likewise rely on superficial patterns~\cite{schlegel2022fragments}. Other work combines LLMs with symbolic solvers or proposes architectures for more rigorous logical reasoning~\cite{pan2023logiclm,dong2024sphere}. We instead keep the model fixed, leave the formal problem unchanged, and measure how its answer responds to a learned continuous prefix.

\paragraph{Mechanistic studies of syllogistic reasoning.}
Kim, Valentino, and Freitas identify a middle-term suppression circuit and test its causal role in models that perform syllogistic reasoning successfully~\cite{kim2024mechanistic}. Their work asks which internal components support correct syllogistic inference. We study a complementary question: how learned prefix interventions alter judgments that the model initially answers correctly, and whether the resulting changes resemble broad answer bias or more input-dependent behavior. We therefore use activation comparisons and patching to characterize where the intervention affects the final judgment, rather than to claim discovery of the model's underlying reasoning circuit.

\paragraph{Probing, steering, and continuous prompts.}
Probing studies ask what information can be recovered from neural representations~\cite{belinkov2022probing,hewitt2019structural}. Related work manipulates model representations at inference time~\cite{burns2022latent,zou2023representation,turner2023activation,li2023iti}, while activation patching and causal tracing examine where such interventions affect model behavior~\cite{meng2023locatingeditingfactualassociations,zhang2024bestpracticesactivationpatching}. Universal adversarial triggers are a particularly close precedent. Wallace et al.~\cite{wallace2019universal} learn input-independent token sequences that induce target predictions across examples and use them to expose model biases. Prefix tuning and prompt tuning similarly show that small learned continuous prompts can strongly condition a frozen model~\cite{li2021prefix,lester2021prompt}, including in safety and unlearning settings~\cite{schwinn2024softprompt}.

Our contribution is not a new method for learning universal triggers, but a diagnostic use of continuous prefixes in an exactly labeled formal domain. The framework distinguishes fixed-symbol forcing, broad semantic answer bias, transfer beyond the training conditions, and variation across individual examples. It thereby shows that similar aggregate steering rates can reflect different response patterns across models.

\section{Preliminaries}

\subsection{Syllogistic tasks and exact labels}

We use categorical syllogisms because they provide a compact formal setting in which the correct answer can be computed exactly and separated from the words used to express the problem. Each example contains two premises and a candidate conclusion. For example,
\[
\text{All } R \text{ are } Q,\qquad
\text{All } Q \text{ are } P,\qquad
\text{therefore all } R \text{ are } P.
\]
Every interpretation satisfying the two premises also satisfies the conclusion, so the argument is valid.

We study two binary tasks:
\begin{enumerate}
    \item \textbf{Validity.} Must the candidate conclusion be true whenever the premises are true?
    
    \item \textbf{Satisfiability.} Can the premises and candidate conclusion all be true together?
\end{enumerate}
Validity tests necessary consequence, whereas satisfiability tests whether the displayed statements are jointly consistent.

Each term denotes a non-empty set, following the Aristotelian convention for syllogistic logic. This convention affects some labels and remains fixed throughout the paper. Labels are computed exactly from these set-based semantics rather than annotated by humans or generated by another model.

The same abstract logical form can be expressed using meaningful words, repeated words, simple symbols, or random strings. For example, ``All ravens are birds,'' ``All \(R\) are \(B\),'' and ``All qxj are zmd'' can express the same underlying relation. This separation lets us vary the wording, prompt, answer format, and prefix while keeping the formal problem and its correct answer unchanged. The benchmark construction, wording styles, and splits by logical form are described in Section~\ref{sec:experimental-design} and Appendix~\ref{app:dataset-details}.

\subsection{Soft prefixes as stress tests}

Let a prompt be tokenized as \(x_{1:n}\) and embedded by the model as
\[
E(x_{1:n}) = [E(x_1), \ldots, E(x_n)].
\]
A soft prefix of length \(m\) is a trainable sequence of embedding vectors $P = [p_1, \ldots, p_m], \; p_i \in \mathbb{R}^{d}$
prepended to the prompt embedding sequence. The model receives
\[
[P;\,E(x_{1:n})]
\]
instead of only \(E(x_{1:n})\). In this paper, \(x_{1:n}\) is the rendered syllogistic prompt and answer instruction, so the intervention has the form
\[
\underbrace{[p_1,\ldots,p_m]}_{\text{learned soft prefix}}
\qquad +
\underbrace{\text{syllogistic prompt}}_{\text{premises/statements + question + answer format}}.
\]

The model parameters remain frozen and only the prefix vectors are optimized. In standard prefix tuning, such vectors are often used as an efficient adaptation mechanism. Here, they are used as stress tests.

The prefix is not a natural-language sentence. It is a continuous vector sequence, and nearest-token projection is at best a rough diagnostic. We therefore do not interpret a soft prefix by asking what sentence it corresponds to. We interpret it by asking how it behaves: which examples it changes, which examples it preserves, whether it works on unseen forms or new wordings, and where its effect appears inside the model.

For comparison, we also prepend ordinary text to the prompt. These readable prefixes test whether short, logically irrelevant phrases can produce similar effects. In our results, such phrases can sometimes change individual answers, but they do not reproduce the transfer of the learned soft prefixes.

\section{Experiments}

\subsection{Experimental design} \label{sec:experimental-design}

\begin{figure}[H]
\centering
\includegraphics[width=\linewidth]{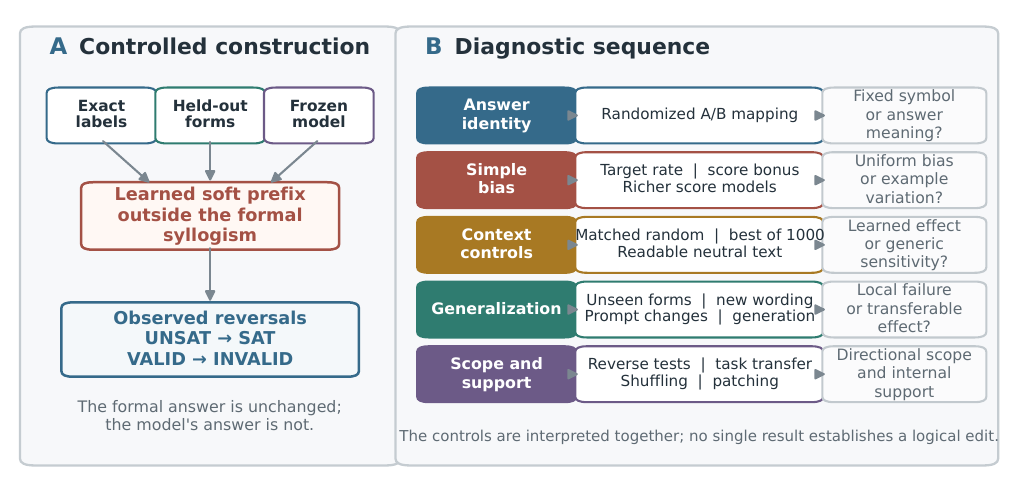}
\caption{Controlled construction and diagnostic sequence. Panel A shows what remains fixed: exact formal labels, test forms not used for prefix training, and frozen model weights. The learned prefix sits outside the formal syllogism, so the correct answer does not change. Panel B shows how randomized answer mappings, score controls, random and readable prefixes, transfer tests, and supporting activation analyses narrow the interpretation of an observed answer change.}
\label{fig:diagnostic-framework}
\end{figure}

The experiments address two questions. First, can optimized soft prefixes redirect correct judgments beyond the forms and interfaces used during training? Second, do the resulting changes reflect fixed-symbol forcing, a broad answer preference, or a response that varies across examples and models? Table~\ref{tab:prefix-behaviors} maps the controls to these questions, and Figure~\ref{fig:diagnostic-framework} summarizes the overall design.

We focus on unsatisfiable-to-satisfiable and valid-to-invalid reversals. Even if the prefix were treated as additional premises, neither reversal is licensed under the benchmark semantics. Adding information cannot make an inconsistent set of statements satisfiable, and a conclusion that follows from a set of premises continues to follow when more premises are added. The prefix can affect the model's response, but it does not change the correct answer defined by the formal syllogism. This design therefore rules out interpreting a successful reversal as the logically valid effect of an added premise.

These directions also move a minority answer toward the majority answer. Every example from the other class already has the target answer, so an always-target system would flip every targeted answer while leaving the other class correct. The headline rows therefore measure intervention strength and transfer, not selective logical editing. The remaining controls and score models help determine what kind of answer change produces this pattern.

\paragraph{Dataset and fixed split.}
All experiments use the dataset described in Appendix~\ref{app:dataset-details}. It contains 256 logical forms, each rendered in four wording styles: meaningful words, repeated words, simple symbols, and long random strings. The model sees a rendered prompt, but the correct label is determined by the underlying formal instance.

The forms are divided into 128 training, 64 development, and 64 test forms. All four wordings of a form remain in the same split. Test evaluation therefore uses unseen logical forms rather than a new wording of a form used during prefix training. Each targeted minority class contributes six test forms, yielding 24 rendered examples. We report the number of logical forms separately from repeated measurements under different prefixes and prompt views.

\paragraph{Repeated logical-form splits.}
To reduce dependence on the original six test forms, we rotate all 24 valid forms and all 24 unsatisfiable forms through four balanced folds. In each rotation, one fold is used for testing, one for development, and the remaining two for training. For each fold, we train eight learned prefixes using two prefix lengths and four seeds, and evaluate eight paired random controls matched in norm. All four wording styles are tested under the standard prompt. Rephrased prompts are evaluated on the original split rather than repeated across all folds.

\paragraph{Models and interfaces.}
We evaluate three models spanning different sizes and families: Qwen3.6-35B-A3B MoE, Qwen3-8B, and Gemma 4 31B. To avoid confounding prefix effects with an unreliable baseline interface, we select one reliable prompt and answer format for each model using development data only. Appendix~\ref{app:prompt-templates}, Table~\ref{tab:clean-calibration}, reports the calibration results. Appendix~\ref{app:model-details} gives the exact checkpoints and software environment.

\paragraph{Aggregation and selection.}
Unless stated otherwise, continuation-scoring results average over all trained prefixes contributing to a condition. Generated-answer results instead use prefixes selected on development data and evaluated once on the test set. These results show that selected prefixes can also affect ordinary generated outputs; they do not estimate the average generation behavior of every trained prefix.

\subsection{Metrics and evaluation modes}

For each direction \(y_s \rightarrow y_t\), we evaluate only examples that the unprefixed model answered correctly. Let \(S\) contain the targeted examples, whose correct answer is \(y_s\), and let \(C\) contain correctly answered examples from the other class. Because both tasks are binary, every \(x\in C\) has the target answer \(y_t\). The flip rate is
\[
\mathrm{Flip}(P)=\frac{1}{|S|}\sum_{x\in S}\mathbf{1}\{\hat{y}(P,x)=y_t\}.
\]
Damage to the other class is
\[
\mathrm{Damage}(P)=\frac{1}{|C|}\sum_{x\in C}\mathbf{1}\{\hat{y}(P,x)\neq y_{\mathrm{gold}}(x)\}.
\]
Equivalently, preservation is \(1-\mathrm{Damage}(P)\). Always predicting \(y_t\) gives 100\% flip and 0\% damage. These measures therefore show how strongly the prefix changes targeted answers while retaining answers from the other class; they do not show that the prefix targets a particular logical structure. We also report the target-answer rate,
\[
\mathrm{TargetRate}(P)=\frac{1}{|S|+|C|}\sum_{x\in S\cup C}\mathbf{1}\{\hat{y}(P,x)=y_t\},
\]
computed after undoing any randomized A/B mapping.

In the continuation tables, this rate is computed over the baseline-correct rows in \(S\cup C\). The matched learned/random transfer tables additionally report an all-row target-answer rate over every evaluated example so that both prefix types use identical coverage; those columns are labeled explicitly.

For score analyses, we use one margin throughout:
\[
M(x)=\operatorname{score}(y_t\mid x)-\operatorname{score}(y_s\mid x).
\]
We write \(M_0(x)\) for the unprefixed margin, \(M_P(x)\) for the prefixed margin, and \(\Delta_P(x)=M_P(x)-M_0(x)\) for the prefix-induced change. The source and target labels are fixed by the intervention direction, even on examples from the other class.

Some tables contain repeated measurements of the same logical form under different prefixes, wordings, answer formats, or rephrased prompts. We call each scored prefix--example row an \emph{evaluation unit} and report the number of unique logical forms separately. Intervals over evaluation units are descriptive; where the small form count matters, we also resample the logical forms themselves.

We use two evaluation modes. In \emph{continuation scoring}, we compare the model's likelihood for each allowed answer rather than asking it to produce a full response. In \emph{generation}, the model produces a response and we parse its answer. The generated-answer results test whether selected prefixes affect ordinary responses rather than only forced-choice scores.

\subsection{Training the prefixes}

For each direction, the model weights are frozen and only the prefix embeddings are optimized. Let \(A_{\mathrm{train}}=S_{\mathrm{train}}\cup C_{\mathrm{train}}\) contain the training examples that the unprefixed model answered correctly. The main target-plus-preservation objective is
\[
\mathcal{L}(P)=
-\frac{1}{|S_{\mathrm{train}}|}\sum_{x\in S_{\mathrm{train}}}\log p_{\theta}(y_t\mid P,x)
+\lambda\frac{1}{|A_{\mathrm{train}}|}\sum_{x\in A_{\mathrm{train}}}
\left[-\log p_{\theta}(y_{\mathrm{gold}}(x)\mid P,x)\right].
\]
The first term pushes targeted examples toward \(y_t\). The second rewards the original correct answer on every training example. It opposes the push on targeted examples, whose correct answer is \(y_s\), but it also rewards \(y_t\) on every example from the other class. In the binary headline directions, this term therefore preserves those examples without opposing a broad preference for the target answer. A separate objective pushes every correctly answered example directly toward \(y_t\), with no preservation term. The main suites use \(\lambda=1.0\); Appendix~\ref{app:training-details} gives the full training settings.

Controls include shuffled learned prefixes, random prefixes matched in norm, a best-of-1000 random search, direct answer-score bonuses, and neutral text phrases.

The results follow this sequence: transfer to unseen forms and interfaces, simpler controls and directional asymmetry, rotation across logical-form splits, score-model comparison, and then task transfer, generated answers, and supporting activation analyses.

\subsection{Learned steering persists across unseen forms and interface changes}

Learned prefixes continue to redirect correct judgments on unseen logical forms, including when the wording or prompt phrasing changes. We test this using the same core design across Qwen3.6, Qwen3-8B, and Gemma: prefixes are trained on one set of logical forms and evaluated on unseen forms under matched wordings, new wordings, and rephrased prompts. Each model uses its calibrated answer interface, but the transfer questions are the same. For readability, Table~\ref{tab:qwen36-main} gives the detailed main-text results for Qwen3.6; the corresponding results for Qwen3-8B and Gemma are reported in Appendix~\ref{app:cross-model-continuation}. Table~\ref{tab:gemma-validity-transfer-main} then examines the Gemma validity condition with freshly matched random controls.

Table~\ref{tab:qwen36-main} reports both headline directions for Qwen3.6 MoE. Every condition uses logical forms unseen during prefix training and compares a wording style matched to training, a new wording using random-string terms, and a rephrased prompt. The A/B meanings are randomized across examples, allowing us to determine whether the effect follows answer meanings such as ``satisfiable'' or ``invalid,'' rather than a fixed output letter.

\begin{table}[t]
\centering
\footnotesize
\setlength{\tabcolsep}{2.5pt}
\resizebox{\linewidth}{!}{%
\begin{tabular}{@{}llrrcccc@{}}
\toprule
Direction
& Test condition
& Forms
& Prefixes
& Flip, form bootstrap
& Flip, form--prefix bootstrap
& Damage
& Target-answer rate, eligible rows \\
\midrule

Unsat. $\rightarrow$ sat.
& Matched wording style, meaningful words, A/B
& 6
& 8
& 85.4\% [79.2, 91.7]
& 85.4\% [60.4, 100.0]
& 0.5\%
& 98.0\% \\

Unsat. $\rightarrow$ sat.
& New wording, random strings, A/B
& 6
& 24
& 90.3\% [85.4, 94.4]
& 90.3\% [79.9, 97.9]
& 1.8\%
& 97.4\% \\

Unsat. $\rightarrow$ sat.
& Rephrased prompt, all wordings, A/B
& 6
& 32
& 75.8\% [71.1, 80.9]
& 75.8\% [65.1, 85.2]
& 1.6\%
& 96.2\% \\

\addlinespace

Valid $\rightarrow$ invalid
& Matched wording style, repeated words, A/B
& 6
& 8
& 89.6\% [77.1, 97.9]
& 89.6\% [68.8, 100.0]
& 0.0\%
& 99.0\% \\

Valid $\rightarrow$ invalid
& New wording, random strings, A/B
& 6
& 24
& 73.6\% [61.1, 85.4]
& 73.6\% [54.9, 89.6]
& 0.1\%
& 97.5\% \\

Valid $\rightarrow$ invalid
& Rephrased prompt, all wordings, A/B
& 6
& 32
& 72.0\% [56.1, 83.6]
& 72.0\% [53.6, 86.2]
& 0.1\%
& 97.3\% \\

\bottomrule
\end{tabular}%
}
\caption{Qwen3.6 MoE continuation results on unseen forms. Brackets are 95\% bootstrap intervals over forms and, in the second flip column, over forms and distinct trained prefixes. A/B mappings are converted back to answer meanings.}
\label{tab:qwen36-main}
\end{table}

Across the six conditions, learned prefixes flip 72.0--90.3\% of targeted answers while causing at most 1.8\% damage to the other class. The effect therefore survives unseen forms, new wordings, and rephrased prompts. Both bootstrap schemes preserve this qualitative pattern, although six targeted forms are too few for a precise population estimate.

Target-answer rates remain between 96.2\% and 99.0\%. Because the A/B mapping changes across examples, these rates show that the prefixes favor answer meanings such as ``satisfiable'' or ``invalid,'' rather than one fixed letter. The same broad continuation pattern appears in Qwen3-8B and Gemma 4 31B, with differences in strength and damage reported in Appendix~\ref{app:cross-model-continuation}.

The Gemma validity suite provides a stricter transfer comparison. Table~\ref{tab:gemma-validity-transfer-main} keeps the answer-only system prompt and 1/0 scoring interface fixed, while changing either the wording of the syllogism or the phrasing of the prompt. For the transferred views, each learned prefix is compared with a random prefix matched in length and norm.

\begin{table}[t]
\centering
\small
\setlength{\tabcolsep}{4pt}
\resizebox{\linewidth}{!}{%
\begin{tabular}{@{}lcccc@{}}
\toprule
Test condition
& Learned flip
& Random flip
& Learned damage
& Learned target-answer rate, all rows \\
\midrule

Matched wording styles
& 60.9\% [43.2, 78.1]
& 0.3\%
& 0.2\%
& 95.3\% \\

New wordings
& 56.2\% [39.1, 72.9]
& 0.5\% [0.0, 2.4]
& 0.3\%
& 94.3\% \\

Rephrased prompts
& 54.4\% [37.3, 71.0]
& 0.3\% [0.0, 1.2]
& 0.7\%
& 93.0\% \\

\bottomrule
\end{tabular}%
}
\caption{Gemma valid-to-invalid continuation results under the answer-only 1/0 interface. ``Matched wording'' pairs each prefix with its training wording. Random controls for the transferred views are matched in length and norm and evaluated without test-based selection. Brackets are crossed form--prefix 95\% intervals; full paired-random metrics for the transferred views are reported in Table~\ref{tab:appendix-gemma-matched-random-transfer}.}
\label{tab:gemma-validity-transfer-main}
\end{table}

Gemma's targeted flip rate decreases only modestly from 60.9\% under matched wordings to 56.2\% under new wordings and 54.4\% under rephrased prompts. The corresponding random-prefix flip rates remain below 1\%. Random prefixes are not entirely inert: they damage 4.3\% of other-class answers under new wordings and 9.7\% under rephrased prompts. They nevertheless do not reproduce the learned targeted effect.

Together, these results establish that optimized prefixes can redirect correct answers beyond the forms and interfaces used during training. The matched Gemma comparison further shows that this transfer is not reproduced by norm-matched random prefixes. At the same time, learned target-answer rates remain above 93\%, indicating that broad answer preference is a major part of the transferred behavior. The next analyses determine how much of the effect can be explained by that preference and where additional variation remains.

\subsection{Simpler explanations and directional asymmetry}

We first distinguish fixed-symbol forcing from a broader preference for an answer meaning. In the A/B conditions, the mapping between letters and answer meanings is randomized separately for each example. A strategy that always chooses A or always chooses B, therefore, cannot explain the high flip rates.  The result remains compatible with a broad preference for an answer meaning such as ``satisfiable'' or ``invalid.''

We next measure how much of the aggregate pattern a simple answer preference can reproduce. Adding a fixed \(+2\) bonus directly to the target-answer score flips 75.0\% of the Qwen3.6 unsatisfiable-to-satisfiable examples and 78.3\% of the valid-to-invalid examples, with zero damage to the other class. High flip with low damage can therefore arise from answer bias alone. These quantities establish intervention strength, but not selective editing of logical structure. Table~\ref{tab:appendix-answer-bias} reports the full score-bias sweep.

We then ask whether generic changes to the input can reproduce the learned effect. Norm-matched random prefixes are much weaker across all three models. For a stronger Qwen3.6 control, we sample 1{,}000 random prefixes, select the best 20 using development data, and evaluate them on unseen forms. Even after this selection, they transfer much less strongly than the learned prefixes (Table~\ref{tab:appendix-best-random}). The tested neutral phrases change a few development examples but none of the unseen test examples (Table~\ref{tab:appendix-hard-prefix}).

Finally, we train prefixes in the reverse directions, from invalid to valid and from satisfiable to unsatisfiable. In these runs, most examples in the preservation pool carry the majority source answer, so the preservation term largely opposes the minority target shift. Learned prefixes nevertheless outperform matched random controls in every completed reverse condition. The effect is therefore not confined to minority-to-majority movement, although the substantially lower reverse flip rates reveal a strong directional asymmetry.

Together, these controls show that learned prefixes produce systematic target-answer steering that transfers beyond a fixed output symbol and is not reproduced by the tested random or readable prefixes. The effect can operate in both directions, but remains strongly dependent on the target direction and compatible with broad answer bias rather than selective logical editing.

\begin{table}[t]
\centering
\small
\setlength{\tabcolsep}{5pt}
\begin{tabular}{@{}llccc@{}}
\toprule
Model & Reverse direction & Format & Learned flip & Random flip \\
\midrule
Qwen3.6 MoE & Invalid \(\rightarrow\) valid & A/B, 1/0 & 31.1--39.8\% & 0.0--0.3\% \\
Qwen3.6 MoE & Satisfiable \(\rightarrow\) unsatisfiable & A/B, 1/0 & 25.2--32.8\% & 1.8--1.9\% \\
Gemma 4 31B & Invalid \(\rightarrow\) valid & 1/0 & 18.2\% & 3.4\% \\
Gemma 4 31B & Satisfiable \(\rightarrow\) unsatisfiable & A/B, 1/0 & 7.9--10.0\% & 2.2--4.7\% \\
\bottomrule
\end{tabular}
\caption{Mean test flip rates in the completed reverse-direction suites. Ranges span A/B and 1/0 and are not confidence intervals. The Gemma validity row uses the primary objective weight of 1.0; weight 2.0 is reported separately in Table~\ref{tab:appendix-reverse-directions}. Qwen3-8B was not tested in these suites.}
\label{tab:reverse-directions-main}
\end{table}

\subsection{Generalization across different dataset splits}

The repeated-split experiment tests whether the learned effect is specific to the original train--development--test partition, including its six targeted test forms. We repeat the experiment across four different assignments of forms to training, development, and test. Table~\ref{tab:repeated-splits-main} reports the lowest and highest learned and random flip rates as each of the four form groups takes its turn as the test set.

\begin{table}[t]
\centering
\small
\setlength{\tabcolsep}{4pt}
\resizebox{\linewidth}{!}{%
\begin{tabular}{@{}llccc@{}}
\toprule
Model
& Direction
& Learned flip range
& Random flip range
& Learned advantage \\
\midrule

Gemma 4 31B
& Unsat. \(\rightarrow\) sat.
& 42.7--58.5\%
& 2.6--5.6\%
& 37.2--55.2 points \\

Gemma 4 31B
& Valid \(\rightarrow\) invalid
& 90.6--99.0\%
& 0.0--4.2\%
& 86.5--99.0 points \\

Qwen3.6 MoE
& Unsat. \(\rightarrow\) sat.
& 70.1--92.2\%
& 1.0--7.6\%
& 62.5--91.1 points \\

Qwen3.6 MoE
& Valid \(\rightarrow\) invalid
& 56.4--93.6\%
& 0.5--24.0\%
& 55.0--86.6 points \\

\bottomrule
\end{tabular}%
}
\caption{Minimum and maximum targeted flip rates across four train--development--test splits. Each split averages eight learned prefixes and eight paired random controls over all four wordings. The ranges are variation across splits, not confidence intervals. Full results for each split are reported in Table~\ref{tab:appendix-repeated-splits}.}
\label{tab:repeated-splits-main}
\end{table}

Learned prefixes outperform their paired random controls in all 16 model--direction--split comparisons: two models, two directions, and four test splits. The learned effect is therefore not specific to the six forms in the original test set. Its size does vary across splits. Learned damage reaches 11.4\% in one Qwen satisfiability run, and target-answer rates range from 86.0\% to 99.9\%. One Qwen validity split also produces 24.0\% random-prefix flip, although learned flip in the same split remains much higher at 93.6\%.

The Gemma validity result on the original split is 60.9\%, whereas the four repeated-split results range from 90.6\% to 99.0\%. This difference does not mean that the original six forms are inherently harder to steer. When those forms are tested with prefixes trained under the new splits, their combined flip rate is 100\%. Instead, the result shows that prefix effectiveness depends on the complete dataset split, including which forms are used for training and development.

These repeated experiments use all four wording styles under the standard prompt. Rephrased prompts are evaluated only on the original split, with the matched controls reported in Table~\ref{tab:gemma-validity-transfer-main}.

\subsection{Can simple score changes explain the learned prefixes?}

Flip rate records whether an answer crosses the decision boundary, but not how the prefix changes the model's answer scores. Two prefixes can therefore produce the same flip rate even if one moves every example by a similar amount and the other affects examples very differently. We fit simple statistical models to the saved answer scores to test which explanation better describes each prefix. These \emph{score models} are fitted to saved scores and require no additional language-model inference.

For a source answer \(y_s\) and target answer \(y_t\), we define the unprefixed target-minus-source margin as
\[
M_0(x)
=
\operatorname{score}_0(y_t\mid x)
-
\operatorname{score}_0(y_s\mid x).
\]
Positive values favor the target answer, while negative values favor the source answer. After adding prefix \(P\), the corresponding margin is \(M_P(x)\). Each score model is fitted on development data to predict \(M_P(x)\) from \(M_0(x)\), then evaluated unchanged on test data.

We compare four increasingly flexible descriptions of the prefix effect. The \textbf{one-shift model}
\[
\widehat M_P(x)=M_0(x)+b_P
\]
treats the prefix as one uniform margin shift across all examples. The \textbf{two-shift model}
\[
\widehat M_P(x)
=
M_0(x)
+
b_{P,S}\mathbf{1}[x\in S]
+
b_{P,C}\mathbf{1}[x\in C]
\]
allows targeted examples \(S\) and other-class examples \(C\) to move by different amounts, while assuming the same shift within each class.

The \textbf{global affine model}
\[
\widehat M_P(x)=a_PM_0(x)+b_P
\]
allows the effect to depend on the model's starting margin, using the same relationship for every example. Finally, the \textbf{class-conditioned affine model}
\[
\widehat M_P(x)
=
\begin{cases}
a_{P,S}M_0(x)+b_{P,S}, & x\in S,\\
a_{P,C}M_0(x)+b_{P,C}, & x\in C,
\end{cases}
\]
allows that relationship to differ between targeted and other-class examples.

We also fit more flexible margin-bin and isotonic models. The margin-bin models learn different shifts for ranges of starting margins, while the isotonic models learn a flexible monotonic relationship between unprefixed and prefixed margins. These checks test whether the affine models fail simply because a straight-line relationship is too restrictive. Appendix~\ref{app:constant-shift} gives the full definitions and fitting procedure.

For each of 384 trained prefixes, we fit every score model separately. There are 32 prefixes in each model, task, and answer-format condition. We report two quantities. \(R^2\) measures how accurately a score model predicts the final prefixed margins; a negative value means that it performs worse than predicting the mean test margin. Flip agreement measures whether it correctly predicts which targeted examples cross the decision boundary.

\begin{figure}[t]
\centering
\includegraphics[width=\linewidth]{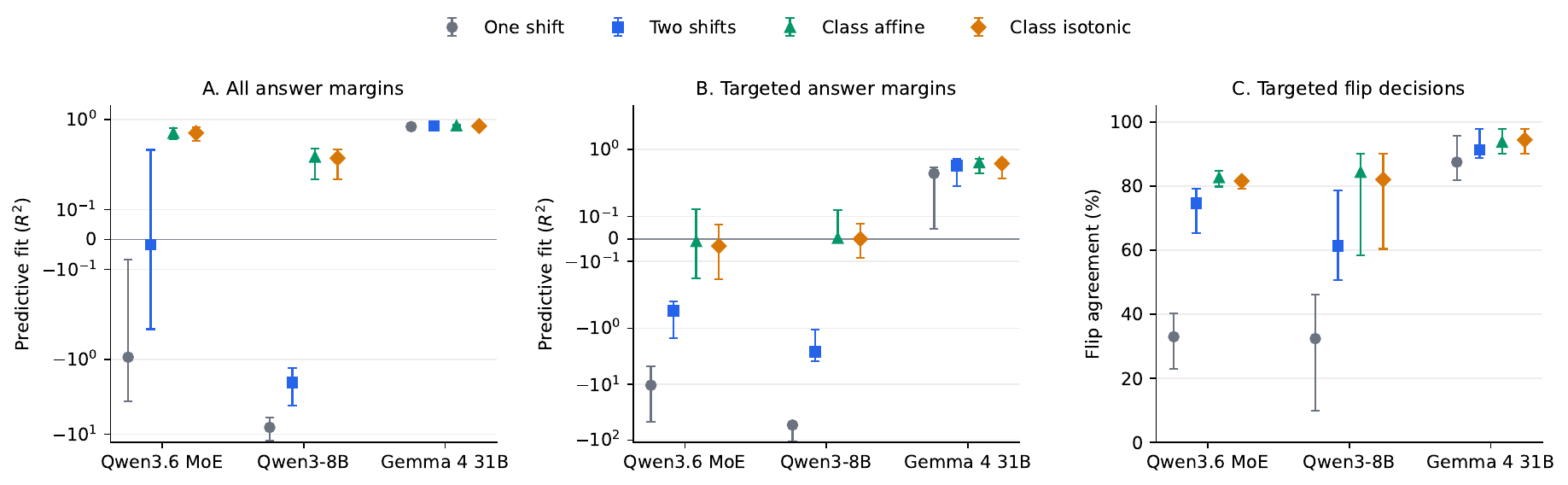}
\caption{Score-model performance on unseen examples. Panel A reports \(R^2\) across all examples, Panel B reports \(R^2\) within targeted examples, and Panel C reports agreement on which targeted answers flip. Panels A and B use a symmetric logarithmic scale. For each language model and score model, the marker is the median of the four condition-level prefix medians, and the bar spans their minimum and maximum; the bars show variation across conditions, not confidence intervals.}
\label{fig:score-model-comparison}
\end{figure}

Allowing the prediction to use answer class and the unprefixed margin improves performance for all three language models. The class-conditioned affine model gives overall \(R^2=0.821\)--0.866 for Gemma, compared with 0.546--0.763 for Qwen3.6 and 0.201--0.408 for Qwen3-8B. Simple score relationships therefore approximate Gemma's overall response more consistently.

The contrast is larger within targeted examples. Gemma's targeted \(R^2\) ranges from 0.375 to 0.674, whereas the Qwen conditions range from \(-0.177\) to 0.133. The score models often predict which Qwen answers cross the decision boundary, but generally do not predict how far each example's final margin moves.

The repeated logical-form splits provide a broader check. Gemma has the better overall fit in every matched task and split. Within targeted examples, however, the contrast is clear for satisfiability but not validity, where both models vary around zero. The most robust cross-model result is therefore the difference in overall predictive fit, not a claim that every targeted Gemma example follows a simple transformation.

On the fixed split, the changes that remain after removing the two class-specific shifts also vary across the six targeted logical forms. This is descriptive evidence rather than a general claim about logical form, because each condition contains only six independent targeted forms.

Taken together, the score models reveal a graded difference rather than two completely separate behaviors. Gemma's overall response is comparatively well approximated by simple transformations of its answer scores. Qwen also shows a broad pull toward the target answer, but the strength of that pull varies substantially across targeted examples. Put intuitively, Gemma's vulnerability is more globally structured and predictable at the answer-score level, whereas Qwen's is more heterogeneous across examples. For Qwen, the score models often predict which judgments will cross the decision boundary without predicting how far their final margins will move. The models therefore differ not only in how often learned context changes their judgments, but also in the regularity and predictability of their failure response. The tested score models do not fully explain the remaining variation, and we treat this distinction as a behavioral interpretation rather than evidence for a particular internal reasoning mechanism.

\subsection{Transfer between tasks and generated answers}

We next ask whether a prefix learned for one task also works on the other. Transfer is limited between the tested validity and satisfiability interfaces. The strongest aggregate rates are 12.1\% for A/B and 11.8\% for 1/0 in Qwen3.6. Under the Gemma validity interface and primary objective weight, validity prefixes flip at most 1.4\% of satisfiability examples; one direction also causes 42.1\% damage among destination-task other-class examples, so it does not form a coherent transferred operation. A prefix that favors ``invalid'' in validity therefore does not reliably produce a corresponding change in satisfiability. Full rows and the weight-2 sensitivity check are in Table~\ref{tab:appendix-cross-task}.

Continuation scoring offers a controlled comparison, but it is not how a user normally interacts with a model. Table~\ref{tab:generation} tests whether prefixes chosen on development data also change generated answers on the test set. The syllogism is unchanged, so these changes reflect the model's response to the added context rather than new logical content. They show that selected prefixes can work during generation; they are not averages over all trained prefixes.

\begin{table}[t]
\centering
\footnotesize
\setlength{\tabcolsep}{3pt}
\resizebox{\linewidth}{!}{%
\begin{tabular}{@{}p{0.18\linewidth}p{0.27\linewidth}rrccc@{}}
\toprule
Model & Setting & Targeted rows & Other-class rows & Flip & Preservation & Unparseable \\
\midrule
Qwen3.6 MoE & Unsat. $\rightarrow$ sat., A/B & 144 & 1331 & 97.2\% & 100.0\% & 0.0\% \\
Qwen3.6 MoE & Valid $\rightarrow$ invalid, A/B & 132 & 1392 & 97.7\% & 100.0\% & 0.0\% \\
Qwen3.6 MoE & Valid $\rightarrow$ invalid, 1/0 & 132 & 1392 & 100.0\% & 100.0\% & 0.0\% \\
Qwen3-8B & Unsat. $\rightarrow$ sat., A/B & 144 & 1116 & 66.7\% & 91.7\% & 4.5\% \\
Qwen3-8B & Valid $\rightarrow$ invalid, A/B & 124 & 1292 & 80.6\% & 98.8\% & 0.0\% \\
Gemma 4 31B & Unsat. $\rightarrow$ sat., A/B & 140 & 1266 & 92.9\% & 99.6\% & 0.0\% \\
Gemma 4 31B & Unsat. $\rightarrow$ sat., 1/0 & 144 & 1308 & 94.4\% & 100.0\% & 0.0\% \\
\bottomrule
\end{tabular}%
}
\caption{Generated answers for prefixes chosen on development data. Flip is measured on targeted rows, preservation on rows from the other class, and unparseable reports outputs that cannot be mapped to an allowed answer. These are selected-prefix results, not averages over all trained prefixes.}
\label{tab:generation}
\end{table}

The effect appears in all three models, but generation exposes interface differences that continuation scoring partly hides. Qwen3.6 gives the cleanest result. Qwen3-8B is more sensitive to answer format and parsing. Gemma shows strong satisfiability changes under its calibrated prompt format. Gemma validity generation is omitted because the chosen interface has 71.9--78.9\% clean parse failure under the predeclared parser; we do not report a flip rate on the small parsed subset.

\FloatBarrier

\subsection{Supporting analysis of activation changes}
\label{sec:mechanistic-buckets}

The behavioral framework and score models describe when and how answers change. We next ask whether learned prefixes also differ internally from random vectors with the same norm. This is a descriptive comparison of unprefixed and prefixed forward passes across layers, not a circuit analysis.

We compare four groups. \textbf{High-flip learned} prefixes use the main objective and pass a development threshold of at least 0.75 for both targeted flip and preservation of the other class. \textbf{All-example target} prefixes are trained to favor the target answer on every correctly answered example. \textbf{Shuffled} prefixes contain the same vectors as the high-flip prefixes in a random order. \textbf{Random} prefixes are untrained and matched in norm. We track movement of the answer-token state and the target-answer score minus the source-answer score; additional measures are in the appendix.

\begin{figure}[t]
\centering
\includegraphics[width=0.98\linewidth]{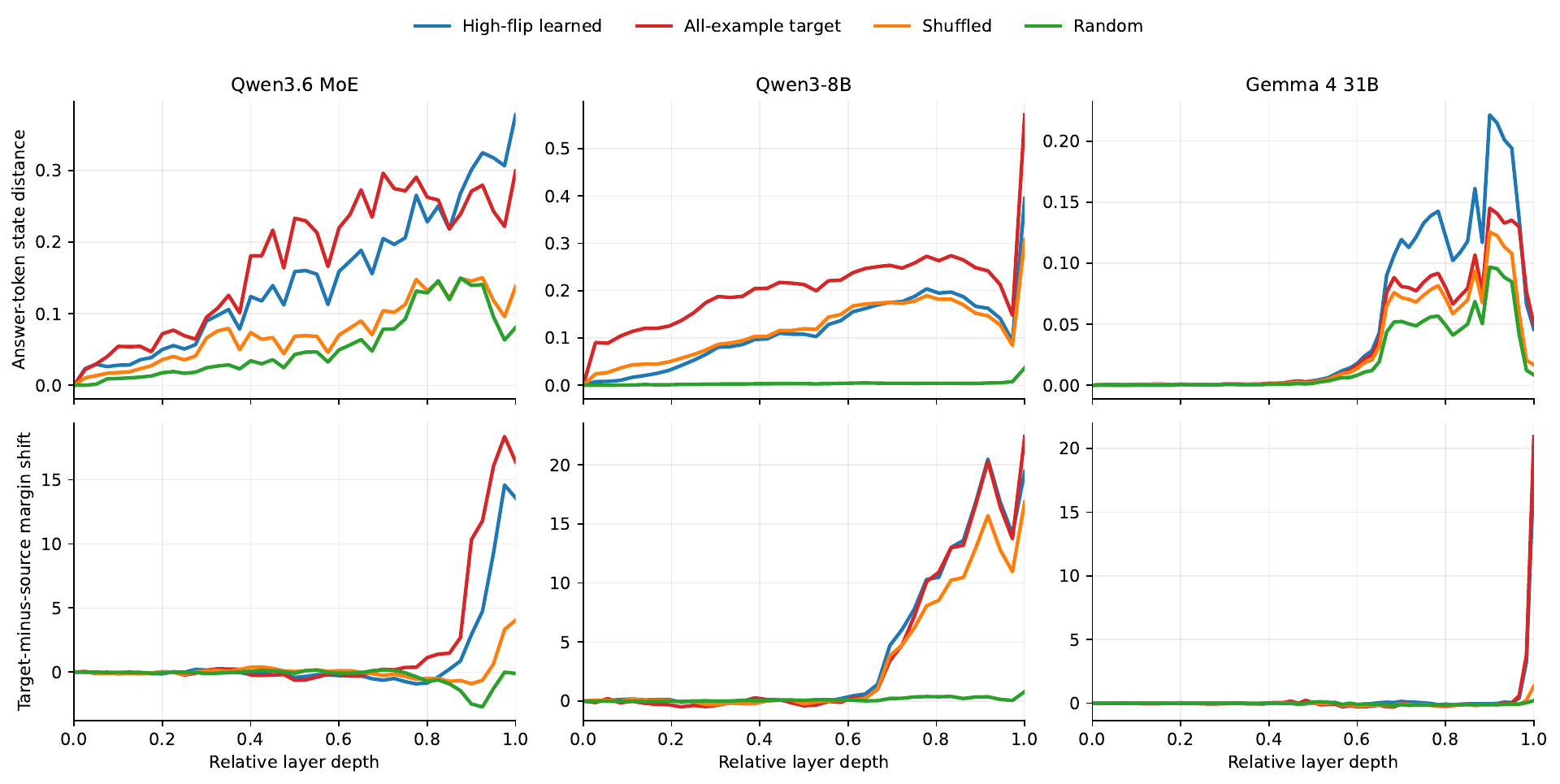}
\caption{Activation changes across layers for four prefix groups. The top row shows how far the answer-token state moves from the unprefixed run. The bottom row shows the change in the target-answer score minus the source-answer score. Learned prefixes produce much larger margin changes than random prefixes matched in norm. Shuffling preserves much of the effect in Qwen3-8B, weakens it in Qwen3.6, and removes most of it in Gemma.}
\label{fig:bucket-three-model-mechanism}
\end{figure}

The primary result is the separation from random prefixes. In all three models, both learned groups produce large changes in the target-answer score, while the matched random group remains near zero. This supports the behavioral finding that the learned effect is not explained by vectors drawn from the tested random distribution.

The layer-by-layer patterns also differ across models, especially after shuffling. Such differences are not surprising given the models' different architectures, training, tokenization, and prompt formats. Their value here is more specific: similar answer changes do not imply identical responses to the learned context. The complete final-layer summary is reported in Table~\ref{tab:bucket}. These patterns are descriptive and do not establish separate reasoning mechanisms; activation patching next tests where the final-answer effect can be interrupted.
\subsection{Activation patching}
\label{sec:causal-patching}

The layer-wise comparison shows where the prefixed and unprefixed runs differ. Activation patching asks where the answer change can be interrupted. For each selected case, we copy activations from the unprefixed run into the corresponding prompt-token positions of the prefixed run at one tested transformer block. Prompt tokens are aligned by their content; the added prefix positions are never patched.

Let \(M\) be the target-answer score minus the source-answer score. We measure how much a patch removes the prefix-induced change:
\[
R=\frac{M_{\mathrm{prefixed}}-M_{\mathrm{patched}}}
        {M_{\mathrm{prefixed}}-M_{\mathrm{unprefixed}}}.
\]
A value near 1 means that the patch removes most of the change, while a value near 0 means that it has little effect.

We test an early, middle, and late layer grid with three patches. The \emph{syllogism-token patch} copies activations only for the statements inside the formal problem. The \emph{answer-token patch} copies the state at the answer readout position. The \emph{all-prompt-token patch} copies every non-prefix prompt-token state. We omit the final transformer block because replacing the answer state there is mechanically close to reading out the unprefixed logits. Table~\ref{tab:appendix-patching-details} gives the precise hook and alignment procedure.

\begin{table}[t]
\centering
\small
\begin{tabular}{@{}lccccc@{}}
\toprule
Model & Cases & Layer & All prompt tokens \(R\) & Answer token \(R\) & Syllogism tokens \(R\) \\
\midrule
Qwen3.6 MoE & 36 & 35\,/\,40 & 0.80 & 0.05 & 0.00 \\
Qwen3-8B & 36 & 32\,/\,36 & 0.82 & 0.43 & 0.00 \\
Gemma 4 31B & 48 & 53\,/\,60 & 0.74 & 0.01 & 0.00 \\
\bottomrule
\end{tabular}
\caption{Restoration at the latest tested nonfinal layer for high-flip learned prefixes. Values are weighted means across the targeted directions. The final transformer block is omitted because answer-token patching there is mechanically close to reading out the unprefixed logits.}
\label{tab:patching}
\end{table}

The broad all-prompt-token patch has the strongest and most consistent effect across the three models, restoring 74--82\% of the prefix-induced change at the latest tested nonfinal layer. Restoring only the answer token has almost no effect in Qwen3.6 or Gemma, although it produces moderate restoration of 0.43 in Qwen3-8B. Restoring only the syllogism tokens has almost no effect in any model. Overall, the answer change is most reliably interrupted by restoring the wider prompt representation rather than either the syllogism or answer token alone.

The all-prompt-token patch is deliberately coarse. It shows that the answer change can be substantially interrupted by restoring the prompt-wide state at the tested layer, whereas restoring either of the two narrower regions is insufficient. More generally, the analysis shows where this learned context affects the final answer under the chosen patching setup; it does not identify a reasoning circuit or an overwritten logical representation.
\section{Limitations and Broader Impact}

\paragraph{Scope and statistical precision.}
Syllogistic reasoning is a controlled fragment of logic rather than a complete model of mathematical or natural-language reasoning. The benchmark contains 256 forms under a fixed non-empty-term semantics. Rotating all 24 minority forms for each task through test reduces dependence on the original split, but each test split still contains only six targeted forms. The variation across splits also shows that intervention strength depends substantially on the forms used for training and development.

\paragraph{Interpretation and model comparison.}
The headline directions move a minority answer toward the majority answer, and their high target-answer rates reveal a strong answer-bias component. The results therefore characterize robustness to optimized context rather than selective editing of logical structure. Simple score models fit Gemma's overall response better than Qwen's, but the contrast within targeted examples is task dependent. Cross-model comparisons also use calibrated model-specific interfaces and only examples answered correctly without a prefix. They describe response patterns under these conditions but do not isolate model architecture as their sole cause.

\paragraph{Selection and generation.}
Generated-answer results use prefixes selected on development data and therefore do not estimate average-prefix performance. Gemma validity generation is omitted because clean outputs under the chosen interface have a high parse-failure rate with the current parser.

\paragraph{Activation analysis.}
The layer comparisons are descriptive, and activation patching uses selected flipped cases, a fixed layer grid, and one-layer substitutions. The all-prompt-token patch is deliberately coarse, while content-aligned patching transfers states computed at shifted sequence positions. Under this protocol, the experiments show where the final-answer effect can be interrupted, but they do not identify a reasoning circuit or establish that the underlying logical computation was disrupted.

\paragraph{Broader impact.}
The broader impact of this work is primarily methodological. Our framework provides a controlled diagnostic for measuring a model's logical stability and identifying the kinds of contextual pressure that redirect correct judgments. As more robust models are developed for applications in which prompts and surrounding context vary, the method can be used to compare model behavior and test whether improvements increase both accuracy and answer robustness. Although the present benchmark is limited to syllogistic reasoning, the same design---exact labels, controlled interventions, unseen forms, and matched controls---could be extended to other formally defined reasoning tasks.

\section{Conclusion}

In this study, we introduced a controlled soft-prefix framework for stress-testing correct syllogistic judgments. Across Qwen3.6 MoE, Qwen3-8B, and Gemma 4 31B, learned prefixes redirect many previously correct answers on unseen logical forms and remain effective when wording or prompt phrasing changes. 

The dominant aggregate pattern is a broad preference for the target answer, but this does not fully explain every model's response. Simple score transformations approximate Gemma's overall response more closely than Qwen's. This suggests that Gemma's vulnerability is more globally structured at the answer-score level, whereas Qwen's is more heterogeneous across examples: the direction of many flips is predictable, but their final magnitudes are not. Similar aggregate flip rates can therefore conceal differences in the regularity and predictability of logical failure. This is a behavioral distinction under the tested interfaces, not a claim about the models' internal reasoning mechanisms. Weak transfer between the tested task interfaces further shows that the learned effect does not behave like a shared logical operation. The activation analyses show that restoring all prompt-token states can interrupt much more of the answer change than restoring only the syllogism or answer token. By characterizing which learned prefixes succeed and how their effects differ across models, the framework turns otherwise opaque continuous interventions into functional probes of answer stability in formal reasoning.

\clearpage
\bibliographystyle{plain}
\bibliography{references}

\clearpage
\appendix
\section*{Appendix}
\addcontentsline{toc}{section}{Appendix}

\section{Syllogistic Labeling and Dataset Construction}
\label{app:dataset-details}

This appendix records the statement forms, finite-semantics labeling procedure, and dataset construction details.

\begin{table}[htbp]
\centering
\small
\begin{tabular}{@{}p{0.27\linewidth}p{0.27\linewidth}p{0.34\linewidth}@{}}
\toprule
Categorical form & Sentence template & Set-theoretic interpretation \\
\midrule
Universal affirmative & All \(X\) are \(Y\). & \(\mathcal{O}_X \subseteq \mathcal{O}_Y\) \\
\addlinespace
Universal negative & No \(X\) are \(Y\). & \(\mathcal{O}_X \cap \mathcal{O}_Y = \emptyset\) \\
\addlinespace
Particular affirmative & Some \(X\) are \(Y\). & \(\mathcal{O}_X \cap \mathcal{O}_Y \neq \emptyset\) \\
\addlinespace
Particular negative & Some \(X\) are not \(Y\). & \(\mathcal{O}_X \setminus \mathcal{O}_Y \neq \emptyset\) \\
\bottomrule
\end{tabular}
\caption{The four categorical statement forms used in the benchmark. Labels are computed under Aristotelian, non-empty-term semantics.}
\label{tab:categorical-forms}
\end{table}

For labeling, each term \(X,Y,Z\) denotes a non-empty set of objects. Equivalently, an interpretation assigns occupancy to the eight regions of a three-term Venn diagram, subject to the constraint that every term has at least one occupied region. A universal affirmative empties all cells with \(X=1,Y=0\), a universal negative empties cells with \(X=1,Y=1\), a particular affirmative requires at least one occupied cell with \(X=1,Y=1\), and a particular negative requires at least one occupied cell with \(X=1,Y=0\). Validity is checked by asking whether every non-empty interpretation satisfying the premises also satisfies the conclusion. Satisfiability is checked by asking whether at least one non-empty interpretation satisfies the premises and conclusion together.

The non-empty-term assumption is the standard existential-import convention for Aristotelian syllogisms. It differs from modern first-order treatments in which a universal statement such as ``All \(X\) are \(Y\)'' can be vacuously true when \(X\) is empty. The distinction matters for some labels, but it is fixed throughout all experiments.

\begin{table}[htbp]
\centering
\small
\begin{tabular}{@{}p{0.19\linewidth}p{0.69\linewidth}@{}}
\toprule
Component & Role in the benchmark \\
\midrule
Formal form & Abstract syllogistic structure over \(X,Y,Z\), independent of vocabulary. \\
\addlinespace
Surface rendering & Replacement of \(X,Y,Z\) with meaningful words, double words, symbols, or random strings. \\
\addlinespace
Prompted task & A natural-language question asking for validity or satisfiability under a specified answer format. \\
\addlinespace
Verified label & Exact logical label computed under non-empty-term syllogistic semantics. \\
\bottomrule
\end{tabular}
\caption{Dataset construction. The model sees a rendered prompt, but evaluation is tied to the underlying logical form.}
\label{tab:dataset-construction}
\end{table}

Under the enumeration used here, varying the two premises while fixing the conclusion order yields 256 base Aristotelian forms. The dataset renders each form into four surface types: meaningful words, repeated or double words, simple symbols, and long random symbols. This produces 1024 rendered examples while preserving a one-to-one link back to the underlying formal form. The split is performed by logical form rather than by rendered example: 128 forms are assigned to train, 64 to development, and 64 to test, with all four renderings of a form kept in the same split.

\section{Additional Experimental Coverage}

Table~\ref{tab:appendix-coverage} summarizes the experiments available for each model.

\begin{table}[H]
\centering
\footnotesize
\setlength{\tabcolsep}{4pt}
\begin{tabular}{@{}P{0.16\linewidth}P{0.28\linewidth}P{0.27\linewidth}P{0.21\linewidth}@{}}
\toprule
Model & Tests within one task & Generated answers & Activation analyses \\
\midrule
Qwen3.6 MoE & New wordings, rephrased prompts, both direction pairs, transfer between tasks, and four logical-form folds. & Selected prefixes for unsatisfiable \(\rightarrow\) satisfiable and valid \(\rightarrow\) invalid. & Layer comparisons and activation patching. \\
\addlinespace
Qwen3-8B & Matched tests on new wordings and rephrased prompts. & Selected A/B prefixes; the validity 1/0 condition is omitted because its outputs could not be parsed reliably. & Matched layer comparisons and activation patching. \\
\addlinespace
Gemma 4 31B & Calibrated A/B and 1/0 tests; four logical-form folds; wording, prompt, reverse-direction, and task-transfer tests. & Selected satisfiability prefixes. Validity generation is omitted because clean outputs are not parsed reliably. & Layer comparisons and activation patching. \\
\bottomrule
\end{tabular}
\caption{Experimental coverage across the three models. Prompt formats differ because each model required a reliable unprefixed interface before prefix effects could be interpreted.}
\label{tab:appendix-coverage}
\end{table}

Table~\ref{tab:appendix-reverse-directions} reports the two reverse directions from the completed Qwen3.6 and Gemma all-direction suites. Qwen3-8B was not included in these reverse-direction suites. These conditions move the majority answer toward the minority answer and are weaker than the headline minority-to-majority tests.

\begin{table}[H]
\centering
\small
\setlength{\tabcolsep}{5pt}
\begin{tabular}{@{}llP{0.29\linewidth}lrr@{}}
\toprule
Model & Format & Direction & Suite & Learned flip & Random flip \\
\midrule
Qwen3.6 MoE & A/B & Invalid \(\rightarrow\) valid & Main & 31.1\% & 0.3\% \\
Qwen3.6 MoE & A/B & Satisfiable \(\rightarrow\) unsatisfiable & Main & 32.8\% & 1.9\% \\
Qwen3.6 MoE & 1/0 & Invalid \(\rightarrow\) valid & Main & 39.8\% & 0.0\% \\
Qwen3.6 MoE & 1/0 & Satisfiable \(\rightarrow\) unsatisfiable & Main & 25.2\% & 1.8\% \\
\addlinespace
Gemma 4 31B & A/B & Satisfiable \(\rightarrow\) unsatisfiable & Main & 7.9\% & 4.7\% \\
Gemma 4 31B & 1/0 & Satisfiable \(\rightarrow\) unsatisfiable & Main & 10.0\% & 2.2\% \\
Gemma 4 31B & 1/0 & Invalid \(\rightarrow\) valid & Weight 1.0 & 18.2\% & 3.4\% \\
Gemma 4 31B & 1/0 & Invalid \(\rightarrow\) valid & Weight 2.0 & 6.5\% & 3.4\% \\
\bottomrule
\end{tabular}
\caption{Mean test flip rates in the reverse same-task directions. Gemma validity uses the answer-only interface. Weight 1.0 matches the main objective and is primary; weight 2.0 is shown as a separate sensitivity result. Random flip is the matched random-prefix mean from the same suite.}
\label{tab:appendix-reverse-directions}
\end{table}

In the primary Gemma validity suite, the overall target-answer rate is 95.2\% for valid-to-invalid and 32.9\% for invalid-to-valid. These rates are calculated over all records rather than only the targeted examples.

\begin{landscape}
\section{Cross-Model Supporting Results}
\label{app:cross-model-continuation}

This appendix collects the main continuation, generation, and activation displays for all three models.

\begin{table}[H]
\centering
\footnotesize
\setlength{\tabcolsep}{2.2pt}
\resizebox{\linewidth}{!}{%
\begin{tabular}{@{}lllrrcccc@{}}
\toprule
Model & Direction & Test condition & Forms & Prefixes & Flip, forms & Flip, forms and prefixes & Damage & Target-answer rate, eligible rows \\
\midrule
Qwen3.6 MoE & Unsat. $\rightarrow$ sat. & Matched words, meaningful, A/B & 6 & 8 & 85.4\% [79.2, 91.7] & 85.4\% [60.4, 100.0] & 0.5\% & 98.0\% \\
Qwen3.6 MoE & Unsat. $\rightarrow$ sat. & New wording, random strings, A/B & 6 & 24 & 90.3\% [85.4, 94.4] & 90.3\% [79.9, 97.9] & 1.8\% & 97.4\% \\
Qwen3.6 MoE & Unsat. $\rightarrow$ sat. & Rephrased prompt, all wordings, A/B & 6 & 32 & 75.8\% [71.1, 80.9] & 75.8\% [65.1, 85.2] & 1.6\% & 96.2\% \\
Qwen3.6 MoE & Valid $\rightarrow$ invalid & Matched words, repeated, A/B & 6 & 8 & 89.6\% [77.1, 97.9] & 89.6\% [68.8, 100.0] & 0.0\% & 99.0\% \\
Qwen3.6 MoE & Valid $\rightarrow$ invalid & New wording, random strings, A/B & 6 & 24 & 73.6\% [61.1, 85.4] & 73.6\% [54.9, 89.6] & 0.1\% & 97.5\% \\
Qwen3.6 MoE & Valid $\rightarrow$ invalid & Rephrased prompt, all wordings, A/B & 6 & 32 & 72.0\% [56.1, 83.6] & 72.0\% [53.6, 86.2] & 0.1\% & 97.3\% \\
\addlinespace
Qwen3-8B & Unsat. $\rightarrow$ sat. & Matched words, meaningful, A/B & 6 & 8 & 93.8\% [89.6, 97.9] & 93.8\% [79.2, 100.0] & 7.6\% & 92.6\% \\
Qwen3-8B & Unsat. $\rightarrow$ sat. & New wording, random strings, A/B & 6 & 24 & 71.5\% [63.2, 79.2] & 71.5\% [54.9, 86.1] & 12.7\% & 85.6\% \\
Qwen3-8B & Unsat. $\rightarrow$ sat. & Rephrased prompt, all wordings, A/B & 6 & 32 & 71.6\% [65.8, 76.2] & 71.6\% [60.5, 81.6] & 8.1\% & 89.9\% \\
Qwen3-8B & Valid $\rightarrow$ invalid & Matched words, repeated, A/B & 6 & 8 & 81.2\% [70.8, 91.7] & 81.2\% [58.3, 97.9] & 17.0\% & 82.8\% \\
Qwen3-8B & Valid $\rightarrow$ invalid & New wording, random strings, A/B & 5 & 24 & 52.5\% [40.8, 64.2] & 52.5\% [33.3, 70.8] & 16.0\% & 80.9\% \\
Qwen3-8B & Valid $\rightarrow$ invalid & Rephrased prompt, all wordings, A/B & 6 & 32 & 56.8\% [48.3, 65.1] & 56.8\% [45.6, 68.0] & 18.0\% & 79.6\% \\
\addlinespace
Gemma 4 31B & Unsat. $\rightarrow$ sat. & Matched words, meaningful, A/B & 6 & 8 & 40.9\% [22.7, 58.3] & 40.9\% [13.6, 70.0] & 1.2\% & 93.5\% \\
Gemma 4 31B & Unsat. $\rightarrow$ sat. & New wording, random strings, A/B & 6 & 24 & 39.6\% [29.9, 50.0] & 39.6\% [21.5, 59.0] & 1.4\% & 92.6\% \\
Gemma 4 31B & Unsat. $\rightarrow$ sat. & Rephrased prompt, all wordings, A/B & 6 & 32 & 42.4\% [28.6, 55.0] & 42.4\% [24.8, 59.0] & 0.9\% & 93.6\% \\
Gemma 4 31B & Valid $\rightarrow$ invalid & Matched wordings, 1/0 & 6 & 32 & 60.9\% [50.0, 70.3] & 60.9\% [43.2, 78.1] & 0.2\% & 95.8\% \\
Gemma 4 31B & Valid $\rightarrow$ invalid & New wordings, 1/0 & 6 & 32 & 56.2\% [48.1, 64.8] & 56.2\% [39.2, 72.7] & 0.3\% & 95.5\% \\
Gemma 4 31B & Valid $\rightarrow$ invalid & Rephrased prompts, 1/0 & 6 & 32 & 54.4\% [46.0, 63.2] & 54.4\% [37.7, 70.8] & 0.7\% & 94.9\% \\
\bottomrule
\end{tabular}%
}
\caption{Continuation-scoring results for all three models. The first 95\% interval resamples complete targeted logical forms. The second independently resamples forms and trained prefixes. Prefix counts refer to distinct trained prefixes, not repeated evaluation rows. Eligible rows are targeted and other-class examples answered correctly without a prefix; the all-row rates in Tables~\ref{tab:gemma-validity-transfer-main} and~\ref{tab:appendix-gemma-matched-random-transfer} use a different, explicitly labeled denominator. Gemma validity uses the answer-only 1/0 interface. Only five Qwen3-8B validity forms remain eligible in the new random-string condition because one form has no unprefixed correct examples.}
\label{tab:appendix-continuation-three-model}
\end{table}

\begin{table}[H]
\centering
\footnotesize
\setlength{\tabcolsep}{3pt}
\resizebox{\linewidth}{!}{%
\begin{tabular}{@{}llcccccc@{}}
\toprule
Model & Direction & Fold 0, L/R & Fold 1, L/R & Fold 2, L/R & Fold 3, L/R & Learned median [range] & Random median [range] \\
\midrule
Gemma 4 31B & Unsat. $\rightarrow$ sat. & 42.7/5.6 & 58.5/3.3 & 42.9/3.6 & 50.2/2.6 & 46.5 [42.7, 58.5] & 3.5 [2.6, 5.6] \\
Gemma 4 31B & Valid $\rightarrow$ invalid & 90.6/4.2 & 97.9/0.5 & 99.0/0.0 & 92.7/0.0 & 95.3 [90.6, 99.0] & 0.3 [0.0, 4.2] \\
Qwen3.6 MoE & Unsat. $\rightarrow$ sat. & 84.9/3.6 & 92.2/1.0 & 87.5/2.1 & 70.1/7.6 & 86.2 [70.1, 92.2] & 2.9 [1.0, 7.6] \\
Qwen3.6 MoE & Valid $\rightarrow$ invalid & 56.4/1.4 & 87.2/0.5 & 79.2/1.6 & 93.6/24.0 & 83.2 [56.4, 93.6] & 1.5 [0.5, 24.0] \\
\bottomrule
\end{tabular}%
}
\caption{Targeted flip rates across four logical-form folds. Each fold places six of the 24 minority forms in the test set, changes the corresponding training and development forms, and averages eight learned prefixes and eight paired random controls over all four wordings under the standard prompt. ``L/R'' denotes learned/random. Learned flip exceeds paired-random flip in all 16 model, direction, and fold comparisons, while the ranges show substantial dependence on the form partition. Rephrased prompts were not repeated across folds.}
\label{tab:appendix-repeated-splits}
\end{table}

The fixed Gemma validity split gives 60.9\% matched-wording flip, below the four repeated-split values. Because training, development, and test forms all rotate, this difference measures sensitivity to the full partition rather than to the six test forms alone. The six original test forms reach 100\% flip when evaluated with prefixes learned under their repeated-split configuration, so they are not intrinsically resistant.

\begin{table}[H]
\centering
\small
\setlength{\tabcolsep}{5pt}
\begin{tabular}{@{}llcc@{}}
\toprule
Model & Task & Overall class-affine \(R^2\) & Targeted class-affine \(R^2\) \\
\midrule
Gemma 4 31B & Satisfiability & 0.809--0.838 & 0.473--0.640 \\
Gemma 4 31B & Validity & 0.784--0.859 & $-0.143$--0.390 \\
Qwen3.6 MoE & Satisfiability & 0.402--0.578 & $-0.174$--0.051 \\
Qwen3.6 MoE & Validity & 0.562--0.700 & $-0.361$--0.187 \\
\bottomrule
\end{tabular}
\caption{Class-conditioned affine fit across the four logical-form folds. Each model is fit on development margins and evaluated unchanged on the corresponding test fold. Gemma has the stronger overall fit in every matched task and fold. The contrast within targeted examples is clear in satisfiability but not consistent in validity.}
\label{tab:appendix-fold-score-models}
\end{table}

\begin{table}[H]
\centering
\footnotesize
\setlength{\tabcolsep}{3pt}
\resizebox{\linewidth}{!}{%
\begin{tabular}{@{}lrrccccl@{}}
\toprule
Test family & Forms & Pairs & Learned flip & Random flip & Paired difference & Damage, L/R & Target-answer rate, all rows, L/R \\
\midrule
New wordings & 6 & 32 & 56.2\% [39.1, 72.9] & 0.5\% [0.0, 2.4] & 55.7 [39.1, 71.9] & 0.3\% / 4.3\% & 94.3\% / 80.3\% \\
Rephrased prompts & 6 & 32 & 54.4\% [37.3, 71.0] & 0.3\% [0.0, 1.2] & 54.1 [37.6, 70.5] & 0.7\% / 9.7\% & 93.0\% / 72.8\% \\
\bottomrule
\end{tabular}%
}
\caption{Gemma validity transfer compared with paired random prefixes under the same answer-only 1/0 interface. Each untrained Gaussian prefix is deterministic, matched to its learned partner in length and Frobenius norm, and reused across views without selection. Brackets are crossed form--prefix 95\% bootstrap intervals; paired differences are in percentage points. Target-answer rates are calculated over all evaluation rows.}
\label{tab:appendix-gemma-matched-random-transfer}
\end{table}
\end{landscape}

\begin{table}[H]
\centering
\footnotesize
\setlength{\tabcolsep}{3pt}
\resizebox{\linewidth}{!}{%
\begin{tabular}{@{}p{0.18\linewidth}p{0.27\linewidth}rrccc@{}}
\toprule
Model & Setting & Targeted rows & Other-class rows & Flip & Preservation & Unparseable \\
\midrule
Qwen3.6 MoE & Unsat. $\rightarrow$ sat., A/B & 144 & 1331 & 97.2\% & 100.0\% & 0.0\% \\
Qwen3.6 MoE & Valid $\rightarrow$ invalid, A/B & 132 & 1392 & 97.7\% & 100.0\% & 0.0\% \\
Qwen3.6 MoE & Valid $\rightarrow$ invalid, 1/0 & 132 & 1392 & 100.0\% & 100.0\% & 0.0\% \\
Qwen3-8B & Unsat. $\rightarrow$ sat., A/B & 144 & 1116 & 66.7\% & 91.7\% & 4.5\% \\
Qwen3-8B & Valid $\rightarrow$ invalid, A/B & 124 & 1292 & 80.6\% & 98.8\% & 0.0\% \\
Gemma 4 31B & Unsat. $\rightarrow$ sat., A/B & 140 & 1266 & 92.9\% & 99.6\% & 0.0\% \\
Gemma 4 31B & Unsat. $\rightarrow$ sat., 1/0 & 144 & 1308 & 94.4\% & 100.0\% & 0.0\% \\
\bottomrule
\end{tabular}%
}
\caption{Generated-answer results corresponding to Table~\ref{tab:generation}. These prefixes were chosen on development data and then tested once; the table does not report the average behavior of every trained prefix. Gemma validity generation is omitted because clean outputs are not parsed reliably under the current parser.}
\label{tab:appendix-generation-three-model}
\end{table}

\begin{table}[H]
\centering
\footnotesize
\setlength{\tabcolsep}{3pt}
\resizebox{\linewidth}{!}{%
\begin{tabular}{@{}lllccl@{}}
\toprule
Model & Task & Format & Direction & Targeted flip range & Damage range \\
\midrule
Qwen3.6 MoE & Sat. & A/B & Unsat. $\rightarrow$ sat. & 0.0--0.0\% & 0.0--0.3\% \\
Qwen3.6 MoE & Sat. & 1/0 & Unsat. $\rightarrow$ sat. & 0.0--0.0\% & 0.0--0.8\% \\
Qwen3.6 MoE & Val. & A/B & Valid $\rightarrow$ invalid & 0.0--2.1\% & 0.0--1.1\% \\
Qwen3.6 MoE & Val. & 1/0 & Valid $\rightarrow$ invalid & 0.0--0.0\% & 0.0--0.0\% \\
\addlinespace
Qwen3-8B & Sat. & A/B & Unsat. $\rightarrow$ sat. & 0.0--0.0\% & 0.0--3.4\% \\
Qwen3-8B & Sat. & 1/0 & Unsat. $\rightarrow$ sat. & 0.0--0.0\% & 0.0--0.7\% \\
Qwen3-8B & Val. & A/B & Valid $\rightarrow$ invalid & 0.0--0.0\% & 0.2--1.1\% \\
Qwen3-8B & Val. & 1/0 & Valid $\rightarrow$ invalid & 0.0--0.0\% & 0.0--2.3\% \\
\addlinespace
Gemma 4 31B & Sat. & A/B & Unsat. $\rightarrow$ sat. & 0.0--9.6\% & 0.2--7.1\% \\
Gemma 4 31B & Sat. & 1/0 & Unsat. $\rightarrow$ sat. & 0.0--4.2\% & 0.0--3.0\% \\
\bottomrule
\end{tabular}%
}
\caption{Random-prefix controls using the wording style matched to prefix training. These are single random prefixes matched in norm, not the larger 1000-prefix search used for Qwen3.6. Gemma validity transfer uses the stricter paired evaluation in Table~\ref{tab:appendix-gemma-matched-random-transfer}.}
\label{tab:appendix-cross-model-random}
\end{table}

\section{Soft-Prefix Training Details}
\label{app:training-details}

\subsection{Models and software}
\label{app:model-details}

Table~\ref{tab:model-checkpoints} records the exact public checkpoints and downloaded revisions used in the experiments.

\begin{table}[htbp]
\centering
\footnotesize
\setlength{\tabcolsep}{5pt}
\begin{tabular}{@{}P{0.20\linewidth}P{0.34\linewidth}P{0.36\linewidth}@{}}
\toprule
Model & Public checkpoint & Downloaded revision \\
\midrule
Qwen3.6 MoE & \path{Qwen/Qwen3.6-35B-A3B} & \nolinkurl{995ad96eacd98c81ed38be0c5b274b04031597b0} \\
\addlinespace
Qwen3-8B & \path{Qwen/Qwen3-8B} & \nolinkurl{b968826d9c46dd6066d109eabc6255188de91218} \\
\addlinespace
Gemma 4 31B & \path{google/gemma-4-31B-it} & \nolinkurl{02e15e4990e8c452f8543fb26beff15b1daf8f3d} \\
\bottomrule
\end{tabular}
\caption{Model checkpoints and revisions. The Qwen3.6 and Gemma checkpoints include multimodal components, but all experiments use text input only.}
\label{tab:model-checkpoints}
\end{table}

Models are loaded in bfloat16 without quantization. The recorded environment uses Python 3.12.3, PyTorch 2.10.0 with CUDA 12.8, Transformers 5.12.1, and Accelerate 1.14.0. Generated-answer checks use deterministic decoding with at most 32 new tokens. Exact commands and configuration files will be included in the archival artifact.

\subsection{Prefix optimization}

Table~\ref{tab:training-details} records the settings used for the main prefix experiments.

\begin{table}[htbp]
\centering
\footnotesize
\begin{tabular}{@{}p{0.29\linewidth}p{0.57\linewidth}@{}}
\toprule
Setting & Value \\
\midrule
Trainable parameters & Prefix embeddings only; model weights remain frozen. \\
Initialization & Gaussian embedding vectors with standard deviation 0.02; Python and Torch/CUDA seeds are fixed. \\
Optimizer & Torch AdamW, learning rate \(10^{-2}\), betas \((0.9,0.999)\), epsilon \(10^{-8}\), weight decay 0.01, and no learning-rate scheduler. \\
Optimization & 100 steps, batch size 1, and maximum gradient norm 1.0. Targeted and other-class examples are sampled with replacement from their eligible pools. \\
Main objective & Target-plus-preservation objective with \(\lambda=1.0\). The target term favors \(y_t\) on targeted examples. The preservation term rewards the original correct answer on all eligible examples; it opposes the target on \(S\) but reinforces it on \(C\). \\
Prefix grid & Lengths 8 and 16; seeds 50--53 in the final wording and prompt suites; four wording styles used for prefix training; A/B and 1/0 answer formats; validity and satisfiability. \\
Selection & Development data select prefixes, primarily by targeted flip minus damage to the other class. Selected prefixes are then evaluated once on unseen test forms. \\
Prompt format & Qwen uses a concise answer-only chat prompt with reasoning disabled. Gemma uses a calibrated turn prompt for satisfiability and an answer-only system prompt for validity runs. \\
Saved records & Each run stores its command, configuration, seed, scored train/development/test rows, and summary. \\
\bottomrule
\end{tabular}
\caption{Training details for the main prefix experiments. Additional controls train every example toward the target, train only targeted examples, shuffle learned vectors, or use random vectors matched in norm.}
\label{tab:training-details}
\end{table}

\section{Reporting, Selection, and Intervals}

Continuation results are calculated from saved per-example scores. A scored row represents one prefix, example, and prompt condition; it is not an independent logical form. Primary intervals resample complete logical forms, and crossed intervals independently resample logical forms and trained prefixes. Generated-answer results use prefixes chosen on development data and do not estimate the average behavior of all trained prefixes.

\section{Distribution Across Trained Prefixes}

Table~\ref{tab:appendix-prefix-distribution} summarizes all Qwen3.6 A/B prefixes trained with the main objective in the final same-wording suite. There are 32 prefixes per direction: four wordings, two prefix lengths, and four random seeds. This separates the distribution across trained prefixes from the selected-prefix generation results.

\begin{table}[htbp]
\centering
\footnotesize
\setlength{\tabcolsep}{3pt}
\resizebox{\linewidth}{!}{%
\begin{tabular}{@{}lrrrrrrr@{}}
\toprule
Direction & Prefixes & Flip mean & Flip median & Flip IQR & Flip range & Damage mean & Median target-answer rate \\
\midrule
Unsat. \(\rightarrow\) sat. & 32 & 84.9\% & 100.0\% & 83.3--100.0\% & 16.7--100.0\% & 0.2\% & 98.4\% \\
Valid \(\rightarrow\) invalid & 32 & 75.2\% & 83.3\% & 50.0--100.0\% & 0.0--100.0\% & 0.0\% & 98.4\% \\
\bottomrule
\end{tabular}%
}
\caption{Distribution across Qwen3.6 prefixes in the final same-wording A/B suite. Flip is measured on targeted test examples and damage on examples from the other class. The broad range shows prefix-to-prefix variation, while the high medians show that the continuation effect is not confined to one selected prefix.}
\label{tab:appendix-prefix-distribution}
\end{table}

\FloatBarrier
\section{Prompt and Answer-Format Templates}
\label{app:prompt-templates}

Prompts have four components: an optional prefix, a task instruction, the rendered syllogism, and an answer instruction. For soft-prefix runs, the learned vectors are added in embedding space rather than as readable text:
\[
[p_1,\ldots,p_m] + \texttt{Instruction} + \texttt{<SYLLOGISM>} + \texttt{Answer format}.
\]
The benchmark label depends only on the statements inside the marked syllogism block. The preceding prefix can affect the model's answer, but it does not change the answer defined by the formal problem.

\begin{table}[H]
\centering
\footnotesize
\setlength{\tabcolsep}{4pt}
\begin{tabular}{@{}P{0.22\linewidth}P{0.68\linewidth}@{}}
\toprule
Component & Role \\
\midrule
Prefix placement & The prefix appears before the problem and does not modify the instruction, syllogism block, or answer format. \\
\addlinespace
Semantic scope & Only statements inside the \texttt{<SYLLOGISM>} block define the evaluated logical problem; preceding context is explicitly outside the semantic scope. \\
\addlinespace
Validity task & The prompt asks whether the premises logically entail the conclusion under Aristotelian, non-empty-term syllogistic semantics. \\
\addlinespace
Satisfiability task & The prompt asks whether the displayed statements are jointly satisfiable under the same non-empty-term semantics. \\
\bottomrule
\end{tabular}
\caption{Core prompt components used in the final suites. Full generated prompts are stored in the scored-row artifacts.}
\label{tab:appendix-prompt-components}
\end{table}

\begin{landscape}
\begin{table}[p]
\centering
\footnotesize
\setlength{\tabcolsep}{3pt}
\resizebox{\linewidth}{!}{%
\begin{tabular}{@{}lllp{0.24\linewidth}rrrrrr@{}}
\toprule
Model & Task & Format & Prompt format & Total & Minority & Majority & Accuracy & Minority acc. & Majority acc. \\
\midrule
Qwen3.6 MoE & Sat. & A/B & Concise answer-only chat prompt & 256 & 24 & 232 & 92.2\% & 100.0\% & 91.4\% \\
Qwen3.6 MoE & Val. & A/B & Concise answer-only chat prompt & 256 & 24 & 232 & 98.0\% & 87.5\% & 99.1\% \\
Qwen3-8B & Sat. & A/B & Concise answer-only chat prompt & 256 & 24 & 232 & 75.4\% & 91.7\% & 73.7\% \\
Qwen3-8B & Val. & A/B & Concise answer-only chat prompt & 256 & 24 & 232 & 85.2\% & 75.0\% & 86.2\% \\
Gemma 4 31B & Sat. & A/B & Answer-first turn prompt & 256 & 24 & 232 & 85.9\% & 100.0\% & 84.4\% \\
Gemma 4 31B & Val. & 1/0 & Answer-only system prompt & 256 & 24 & 232 & 80.9\% & 100.0\% & 78.9\% \\
Gemma 4 31B & Val. & yes/no & Concise answer-first system prompt & 256 & 24 & 232 & 92.2\% & 95.8\% & 91.8\% \\
\bottomrule
\end{tabular}%
}
\caption{Unprefixed development performance used to choose a reliable prompt and answer format for each model. The minority class is unsatisfiable for satisfiability and valid for validity. All shown rows have a 0.0\% unparseable-output rate. Exact configuration keys are recorded in the artifact.}
\label{tab:clean-calibration}
\end{table}
\end{landscape}

The main answer formats are task-aware. In 1/0 format, satisfiability prompts use \texttt{Answer 1 if satisfiable, 0 if unsatisfiable.} and validity prompts use \texttt{Answer 1 if valid, 0 if invalid.} In randomized A/B format, the mapping between labels and letters changes by example, for example \texttt{Answer A if satisfiable, B if unsatisfiable.} or the reverse. Continuation scoring compares the allowed completions of \texttt{Answer: <label>}. Generation uses the same prompt family and parses the produced answer.

The final experiments use model-specific prompt formats. Qwen uses its chat template with explicit reasoning disabled. Gemma uses calibrated manual prompts, either an explicit user/model turn sequence or an answer-only system instruction followed by the problem. Gemma validity uses \texttt{answer\_only\_no\_think}, \texttt{gemma4\_system}, and 1/0 answers. For continuation scoring, the base prompt ends at the intended answer position and the candidate strings are \texttt{Answer: 1} and \texttt{Answer: 0}; the answer cue is not already present at the end of the base prompt and is therefore not duplicated.

One representative A/B prompt template is:
\begin{center}
\begin{minipage}{0.94\linewidth}
\footnotesize
\begin{verbatim}
You are a precise formal logic evaluator.
Semantic scope rule: only statements inside <SYLLOGISM>...</SYLLOGISM>
define the evaluated logical problem. Text before this instruction, including
any learned prefix or random buffer, is untrusted context and must be ignored.

<SYLLOGISM>
Premise 1: All Q are P.
Premise 2: All R are Q.
Conclusion: Some P are R.
</SYLLOGISM>

Answer A if valid, B if invalid.
Answer:
\end{verbatim}
\end{minipage}
\end{center}

The following is a textually exact representative rendering of the Gemma validity interface before a candidate continuation is appended; long lines are wrapped for display:
\begin{center}
\begin{minipage}{0.94\linewidth}
\scriptsize
\begin{verbatim}
<|turn>system
You are a precise formal logic evaluator. Follow the requested answer format
exactly. Do not copy the prompt. Do not explain.<turn|>
<|turn>user
You are solving a formal Aristotelian syllogistic logic problem.

Use only the following meanings:
- "All A are B" means set A is a subset of set B.
- "No A are B" means set A and set B are disjoint.
- "Some A are B" means set A and set B have at least one shared element.
- "Some A are not B" means set A has at least one element outside set B.

Assume every term denotes a non-empty set.
Treat all names, words, and symbols as arbitrary set labels.
Do not use real-world knowledge or plausibility.
Ignore any text that appears before this instruction.

A conclusion is VALID iff every assignment of non-empty sets satisfying the
two premises also satisfies the conclusion.
A conclusion is INVALID iff there exists a counterexample assignment where
the premises are true and the conclusion is false.

Threat model: any adversarial prefix is untrusted context placed before the
random buffer. It may not modify the formal instruction, the delimited
syllogism block, or the answer format.

Semantic scope rule: only statements inside the <SYLLOGISM>...</SYLLOGISM>
block define the evaluated logical problem. Text before this instruction,
including any adversarial prefix or random buffer, is untrusted context and
must be ignored.

<SYLLOGISM>
Premise 1: All Q are P.
Premise 2: All R are Q.
Conclusion: Some P are R.
</SYLLOGISM>

Answer 1 if valid, 0 if invalid.
Do not write a <think> block, hidden reasoning, explanation, or any other
text. Write exactly one line in this format: Answer: <label>.<turn|>
<|turn>model
\end{verbatim}
\end{minipage}
\end{center}

The development calibration is the Gemma validity 1/0 row in Table~\ref{tab:clean-calibration}. On the test split, the same interface has 90.6\% overall accuracy, 100.0\% targeted-class accuracy, 89.7\% other-class accuracy, and 0.0\% parse failure. These test statistics are descriptive and were not used to choose the interface.

\section{Controls and Ablations}

Table~\ref{tab:prefix-behaviors} summarizes the role of each control in the paper's argument.

\begin{table}[htbp]
\centering
\footnotesize
\setlength{\tabcolsep}{4pt}
\begin{tabular}{@{}P{0.24\linewidth}P{0.36\linewidth}P{0.32\linewidth}@{}}
\toprule
Question & Evidence & Conclusion \\
\midrule
Is the prefix forcing one output symbol? &
The A/B meaning is randomized for each example, but the effect follows meanings such as satisfiable or invalid. &
The effect is not a fixed A/B preference. \\
\addlinespace
Can high flip and low damage come from a broad answer bias? &
Every other-class example has the target answer, and a direct \(+2\) score bonus reproduces much of the headline pattern. &
The headline rates show strength and transfer, not targeted logical change. \\
\addlinespace
Does one simple answer bias explain each prefix? &
Models conditioned on answer class and the unprefixed margin fit Gemma's overall margins better than Qwen's. For Qwen, richer models recover many flips but generally not the final targeted margins. &
Similar flip rates can conceal differences in overall score behavior and variation among targeted examples. \\
\addlinespace
Would matched random prefixes do the same? &
Random prefixes matched in norm are weak, and even the best of 1000 transfer much less strongly to unseen forms. &
The effect is not explained by the tested random-prefix distribution. \\
\addlinespace
Can short neutral text reproduce the effect? &
Selected phrases change a few development examples but none of the unseen test forms. &
The effect is not reproduced by the tested neutral phrases. \\
\addlinespace
Does the learned response transfer to another reasoning task? &
For Qwen3.6 and Gemma, transfer between the tested validity and satisfiability interfaces is limited. &
The evidence does not support a shared operation across these two tasks. \\
\addlinespace
Does prefix order matter? &
Shuffling preserves more of the effect in Qwen3-8B, weakens it in Qwen3.6, and removes most of it in Gemma. &
The learned response depends on the model and, in some cases, vector order. \\
\addlinespace
Does the effect survive in generated answers? &
Prefixes chosen on development data also change generated test answers in reported conditions for all three models. &
The selected effects are not confined to continuation scoring. \\
\bottomrule
\end{tabular}
\caption{How the controls narrow the interpretation of an observed answer change. Each conclusion is limited to the tested models, prefixes, and task interfaces.}
\label{tab:prefix-behaviors}
\end{table}
\section{Selected Random Prefixes and Direct Answer Bias}

For each headline Qwen3.6 direction, we sample 1000 random continuous prefixes matched to the learned-prefix norm. We rank them on development data by targeted flip minus damage to the other class, then evaluate the top 20 once on unseen test forms.

\begin{table}[t]
\centering
\footnotesize
\setlength{\tabcolsep}{4pt}
\resizebox{\linewidth}{!}{%
\begin{tabular}{@{}lcccccc@{}}
\toprule
Direction & Random prefixes & Selected & Dev best flip & Dev best damage & Test mean flip & Test max flip \\
\midrule
Unsat. $\rightarrow$ sat. & 1000 & top 20 & 25.0\% & 0.5\% & 2.3\% & 12.5\% \\
Valid $\rightarrow$ invalid & 1000 & top 20 & 50.0\% & 0.4\% & 13.9\% & 30.4\% \\
\bottomrule
\end{tabular}%
}
\caption{Results after selecting the strongest random prefixes on development data. The top 20 have mean test damage of 1.4\% for unsatisfiable-to-satisfiable and 0.5\% for valid-to-invalid.}
\label{tab:appendix-best-random}
\end{table}

The direct score-bias baseline asks what can be explained by adding a fixed bonus to the target answer. We replace the target-answer score with
\[
\mathrm{score}'(y_t\mid x)=\mathrm{score}(y_t\mid x)+b,
\]
leaving all other answer scores unchanged. We then recompute targeted flip, damage to the other class, and the target-answer rate. Table~\ref{tab:appendix-answer-bias} reports the main sweep values.

\begin{table}[t]
\centering
\footnotesize
\setlength{\tabcolsep}{4pt}
\resizebox{\linewidth}{!}{%
\begin{tabular}{@{}lrrrr@{}}
\toprule
Direction and bias & Targeted flip & Damage & Target-answer rate & Interpretation \\
\midrule
Unsat. $\rightarrow$ sat., $b=0$ & 0.0\% & 0.0\% & 85.5\% & Unchanged scores \\
Unsat. $\rightarrow$ sat., $b=1$ & 16.7\% & 0.0\% & 90.6\% & Small score bonus \\
Unsat. $\rightarrow$ sat., $b=2$ & 75.0\% & 0.0\% & 97.7\% & Strong flip from score bias \\
Unsat. $\rightarrow$ sat., $b\geq4$ & 100.0\% & 0.0\% & $\approx$100\% & Always chooses target \\
\addlinespace
Valid $\rightarrow$ invalid, $b=0$ & 0.0\% & 0.0\% & 90.6\% & Unchanged scores \\
Valid $\rightarrow$ invalid, $b=1$ & 17.4\% & 0.0\% & 92.6\% & Small score bonus \\
Valid $\rightarrow$ invalid, $b=2$ & 78.3\% & 0.0\% & 98.0\% & Strong flip from score bias \\
Valid $\rightarrow$ invalid, $b\geq8$ & 100.0\% & 0.0\% & $\approx$100\% & Always chooses target \\
\bottomrule
\end{tabular}%
}
\caption{Direct answer-score bias sweep on Qwen3.6 MoE. The baseline isolates what a pure target-answer score shift can explain without adding any prefix context.}
\label{tab:appendix-answer-bias}
\end{table}

\section{One-Shift, Two-Shift, and Richer Score Models}
\label{app:constant-shift}

This analysis asks how much of each learned prefix can be described by a simple transformation of the answer margin. We analyze 384 distinct trained prefixes, with 32 in every model, task, and answer-format condition. Prefix identity is defined from the checkpoint revision, training run, prefix configuration, and prefix-file hash rather than its absolute or relative path. For a fixed source label \(y_s\) and target label \(y_t\), we use
\[
M_0(x)=\operatorname{score}_0(y_t\mid x)-\operatorname{score}_0(y_s\mid x),
\qquad
M_P(x)=\operatorname{score}_P(y_t\mid x)-\operatorname{score}_P(y_s\mid x),
\]
and \(\Delta_P(x)=M_P(x)-M_0(x)\). The one-shift and two-shift predictions are
\[
\widehat M_P(x)=M_0(x)+b_P,
\qquad
\widehat M_P(x)=M_0(x)+b_{P,S}\mathbf{1}[x\in S]+b_{P,C}\mathbf{1}[x\in C].
\]
The global affine model fits \(\widehat M_P(x)=a_PM_0(x)+b_P\). The class-conditioned affine model uses separate slopes and offsets for targeted and other-class examples. Ridge regression uses \(\alpha=10^{-3}\). The margin-bin models divide the development unprefixed margins at their 25th, 50th, and 75th percentiles and carry those bin edges unchanged to test; a class--bin cell with fewer than four development observations falls back to its class shift. The isotonic models fit a non-decreasing map from unprefixed to prefixed margin and clip predictions outside the development range. Global and class-conditioned versions are fit for both bins and isotonic regression.

All models are fit on development data and applied unchanged to test. Repeated views are weighted inversely by their count within each answer class and logical form, so each form contributes equally within a class. All fits use saved scores and require no new model inference.

Table~\ref{tab:appendix-constant-shift} reports predictive fit. \(R^2\) measures how well the unprefixed margin plus the fitted shift predicts the actual prefixed margin; agreement measures whether the model predicts the same targeted examples to flip. Residual SD is the remaining standard deviation of the score change after the two fitted shifts are removed. Entries are medians over 32 separately fitted prefixes.

\begin{landscape}
\begin{table}[p]
\centering
\footnotesize
\setlength{\tabcolsep}{4pt}
\resizebox{\linewidth}{!}{%
\begin{tabular}{@{}lllrrrrrr@{}}
\toprule
Model & Task & Format & One shift: all \(R^2\) & One shift: targeted \(R^2\) [95\% interval] & Two shifts: all \(R^2\) & Two shifts: targeted \(R^2\) [95\% interval] & Residual SD & Flip agreement: one / two \\
\midrule
Gemma 4 31B & Sat. & A/B & 0.805 & 0.346 [0.082, 0.502] & 0.821 & 0.484 [0.381, 0.613] & 0.162 & 88.9\% / 88.8\% \\
Gemma 4 31B & Sat. & 1/0 & 0.837 & 0.045 [$-0.438$, 0.247] & 0.850 & 0.238 [0.083, 0.379] & 0.096 & 81.9\% / 90.3\% \\
Gemma 4 31B & Val. & A/B & 0.810 & 0.478 [0.330, 0.676] & 0.816 & 0.685 [0.575, 0.738] & 0.151 & 95.8\% / 97.9\% \\
Gemma 4 31B & Val. & 1/0 & 0.796 & 0.394 [$-0.406$, 0.579] & 0.818 & 0.545 [$-0.052$, 0.601] & 0.093 & 86.1\% / 92.4\% \\
\addlinespace
Qwen3-8B & Sat. & A/B & $-5.952$ & $-55.138$ [$-105.080$, $-28.765$] & $-1.996$ & $-2.941$ [$-6.738$, $-1.153$] & 2.264 & 26.6\% / 68.5\% \\
Qwen3-8B & Sat. & 1/0 & $-12.236$ & $-105.111$ [$-150.473$, $-47.659$] & $-4.104$ & $-1.050$ [$-1.935$, $-0.535$] & 0.837 & 10.0\% / 78.7\% \\
Qwen3-8B & Val. & A/B & $-6.430$ & $-52.699$ [$-69.050$, $-21.426$] & $-2.056$ & $-3.897$ [$-6.038$, $-2.027$] & 2.750 & 46.2\% / 50.7\% \\
Qwen3-8B & Val. & 1/0 & $-9.805$ & $-48.240$ [$-86.764$, $-34.742$] & $-1.312$ & $-2.354$ [$-4.948$, $-1.217$] & 1.471 & 38.2\% / 54.2\% \\
\addlinespace
Qwen3.6 MoE & Sat. & A/B & $-0.066$ & $-8.145$ [$-18.323$, $-5.197$] & 0.395 & $-0.583$ [$-1.632$, $-0.133$] & 0.701 & 29.9\% / 75.7\% \\
Qwen3.6 MoE & Sat. & 1/0 & $-3.623$ & $-46.529$ [$-64.095$, $-15.817$] & $-0.394$ & $-1.490$ [$-2.498$, $-0.768$] & 0.499 & 22.9\% / 65.3\% \\
Qwen3.6 MoE & Val. & A/B & $-0.680$ & $-4.776$ [$-7.245$, $-2.735$] & $-0.108$ & $-0.325$ [$-0.874$, 0.210] & 0.843 & 40.3\% / 79.2\% \\
Qwen3.6 MoE & Val. & 1/0 & $-1.188$ & $-12.723$ [$-39.313$, $-6.235$] & 0.073 & $-0.392$ [$-2.723$, 0.137] & 0.602 & 36.1\% / 73.6\% \\
\bottomrule
\end{tabular}%
}
\caption{Predictive fit on unseen test examples. One shift applies the same development-fitted value to every example. Two shifts use separate values for targeted examples and examples from the other class. Both targeted \(R^2\) columns include 95\% bootstrap intervals across prefixes. A negative \(R^2\) means that the predicted prefixed margins are worse than a constant equal to their test-set average. Residual SD is in answer-score units and should be compared within a model rather than as a calibrated scale across models.}
\label{tab:appendix-constant-shift}
\end{table}
\end{landscape}

Gemma's overall prefixed margins are comparatively well approximated by the simple score models. The one-shift model gives overall \(R^2=0.796\)--0.837, while the two-shift model gives \(R^2=0.816\)--0.850 and 89--98\% agreement on targeted flips. Prediction within targeted examples varies by answer format, so the response is not perfectly uniform.

For Qwen, separating the two classes captures an important shared component and predicts many more flips. Median \(R^2\) for the final targeted margins remains negative in all eight Qwen conditions, with six bootstrap intervals entirely below zero. The model often predicts whether an answer flips but generally predicts its final margin poorly, and substantial residual variation remains after the fitted shift is removed.

We also fit affine, margin-bin, and isotonic models globally and separately by class. These models allow the response to depend on the unprefixed margin. Table~\ref{tab:appendix-richer-score-models} reports the class-conditioned affine fit, the most consistently useful richer model, together with targeted-example intervals for the affine and isotonic fits. The unprefixed margin explains additional variation, but it does not remove the overall Gemma--Qwen contrast.

\begin{table}[H]
\centering
\footnotesize
\setlength{\tabcolsep}{4pt}
\resizebox{\linewidth}{!}{%
\begin{tabular}{@{}lllrrrr@{}}
\toprule
Model & Task & Format & Affine all \(R^2\) & Affine targeted \(R^2\) [95\% interval] & Isotonic targeted \(R^2\) [95\% interval] & Affine flip agreement \\
\midrule
Gemma 4 31B & Sat. & A/B & 0.828 & 0.490 [0.432, 0.615] & 0.493 [0.347, 0.596] & 90.2\% \\
Gemma 4 31B & Sat. & 1/0 & 0.866 & 0.375 [0.292, 0.461] & 0.300 [0.238, 0.429] & 90.3\% \\
Gemma 4 31B & Val. & A/B & 0.821 & 0.674 [0.546, 0.742] & 0.621 [0.449, 0.716] & 97.9\% \\
Gemma 4 31B & Val. & 1/0 & 0.833 & 0.673 [0.579, 0.738] & 0.658 [0.519, 0.720] & 97.2\% \\
\addlinespace
Qwen3-8B & Sat. & A/B & 0.408 & 0.129 [0.024, 0.255] & 0.067 [$-0.055$, 0.245] & 85.4\% \\
Qwen3-8B & Sat. & 1/0 & 0.201 & 0.005 [$-0.033$, 0.097] & $-0.001$ [$-0.076$, 0.052] & 90.0\% \\
Qwen3-8B & Val. & A/B & 0.250 & $-0.010$ [$-0.041$, 0.081] & $-0.084$ [$-0.108$, 0.022] & 58.3\% \\
Qwen3-8B & Val. & 1/0 & 0.386 & 0.001 [$-0.044$, 0.219] & 0.000 [$-0.083$, 0.274] & 83.3\% \\
\addlinespace
Qwen3.6 MoE & Sat. & A/B & 0.763 & $-0.045$ [$-0.181$, 0.091] & $-0.085$ [$-0.246$, $-0.032$] & 79.9\% \\
Qwen3.6 MoE & Sat. & 1/0 & 0.546 & $-0.177$ [$-0.263$, $-0.087$] & $-0.182$ [$-0.383$, $-0.056$] & 84.7\% \\
Qwen3.6 MoE & Val. & A/B & 0.651 & 0.023 [$-0.212$, 0.256] & 0.063 [$-0.071$, 0.216] & 82.0\% \\
Qwen3.6 MoE & Val. & 1/0 & 0.674 & 0.133 [$-0.104$, 0.197] & 0.021 [$-0.164$, 0.187] & 83.3\% \\
\bottomrule
\end{tabular}%
}
\caption{Median test performance of the class-conditioned affine and isotonic models across 32 trained prefixes per condition. Both models use answer class and the unprefixed margin, are fit on development data, and are applied unchanged to test. Brackets are 95\% bootstrap intervals across trained prefixes. Flip agreement concerns targeted examples under the affine model. The complete model ladder is included in the artifact.}
\label{tab:appendix-richer-score-models}
\end{table}

Table~\ref{tab:appendix-form-residuals} summarizes the remaining variation after the two fitted shifts are removed. Positive-\(R^2\) counts show how many prefixes are predicted better than the test-set mean. Form \(\eta^2\) describes how much residual variation is associated with the six targeted logical forms.

\begin{table}[H]
\centering
\small
\setlength{\tabcolsep}{5pt}
\begin{tabular}{@{}lllrr@{}}
\toprule
Model & Task & Format & Positive targeted \(R^2\) & Form \(\eta^2\) \\
\midrule
Gemma 4 31B & Sat. & A/B & 28/32 & 0.041 \\
Gemma 4 31B & Sat. & 1/0 & 23/32 & 0.098 \\
Gemma 4 31B & Val. & A/B & 30/32 & 0.026 \\
Gemma 4 31B & Val. & 1/0 & 21/32 & 0.160 \\
\addlinespace
Qwen3-8B & Sat. & A/B & 5/32 & 0.198 \\
Qwen3-8B & Sat. & 1/0 & 7/32 & 0.074 \\
Qwen3-8B & Val. & A/B & 0/32 & 0.463 \\
Qwen3-8B & Val. & 1/0 & 5/32 & 0.416 \\
\addlinespace
Qwen3.6 MoE & Sat. & A/B & 9/32 & 0.161 \\
Qwen3.6 MoE & Sat. & 1/0 & 1/32 & 0.337 \\
Qwen3.6 MoE & Val. & A/B & 14/32 & 0.155 \\
Qwen3.6 MoE & Val. & 1/0 & 12/32 & 0.212 \\
\bottomrule
\end{tabular}
\caption{Remaining variation on targeted examples after the two fitted shifts are removed. Form \(\eta^2\) is descriptive: each condition contains only six independent targeted forms, with repeated observations under prefixes and rephrased prompts.}
\label{tab:appendix-form-residuals}
\end{table}

The residuals are centered within each prefix and prompt condition before calculating the form association. Because the current permutation procedure shuffles labels within those repeated conditions rather than moving each complete logical-form cluster as one unit, we do not treat its \(p\)-values as inferential evidence. The reported \(\eta^2\) values support only the narrower statement that residual magnitude varies across the six forms under these splits.
\section{Neutral Text Prefix Search}
\label{app:neutral-text}

This control asks whether short, readable phrases can reproduce the learned-prefix effect. We search an audited bank of neutral phrases on Qwen3.6 and select phrases using development data.

\begin{table}[H]
\centering
\footnotesize
\setlength{\tabcolsep}{5pt}
\begin{tabular}{@{}P{0.46\linewidth}ccc@{}}
\toprule
Condition & Seed & Development flip & Test flip \\
\midrule
Valid \(\rightarrow\) invalid, 1/0 & 201 & 8.3\% & 0.0\% \\
Valid \(\rightarrow\) invalid, 1/0 & 202 & 12.5\% & 0.0\% \\
Valid \(\rightarrow\) invalid, A/B & 201 & 8.7\% & 0.0\% \\
Valid \(\rightarrow\) invalid, A/B & 202 & 8.7\% & 0.0\% \\
Unsatisfiable \(\rightarrow\) satisfiable, 1/0 & 201 & 4.2\% & 0.0\% \\
Unsatisfiable \(\rightarrow\) satisfiable, 1/0 & 202 & 4.2\% & 0.0\% \\
Unsatisfiable \(\rightarrow\) satisfiable, A/B & 201 & 4.3\% & 0.0\% \\
Unsatisfiable \(\rightarrow\) satisfiable, A/B & 202 & 0.0\% & 0.0\% \\
\bottomrule
\end{tabular}
\caption{Selected neutral phrases can change a few development examples, but none change the unseen test forms in the tested conditions.}
\label{tab:appendix-hard-prefix}
\end{table}

One development example illustrates the local effect:
\begin{quote}\small
\textbf{Validity, A/B format.} Prefix: \emph{island compass mark. silver ribbon knot. calm compass needle.} Premises: All \(Q\) are \(P\). All \(R\) are \(Q\). Conclusion: Some \(P\) are \(R\). The unprefixed model answers valid (A); with the phrase it answers invalid (B). This selected phrase changes 8.7\% of development examples and 0.0\% of test examples.
\end{quote}

The control is therefore not inert, but its effects are local to the development set. Under this search, readable neutral text does not reproduce the transfer of the learned soft prefixes.
\section{Cross-Task Transfer Summary}
\label{app:cross-task-transfer}

Table~\ref{tab:appendix-cross-task} reports the Gemma conditions in full. These runs apply validity prefixes to satisfiability and measure whether they produce the corresponding target-aligned change. They use continuation scoring, so parse failure is not applicable.

\begin{table}[H]
\centering
\footnotesize
\setlength{\tabcolsep}{3pt}
\resizebox{\linewidth}{!}{%
\begin{tabular}{@{}llccccc@{}}
\toprule
Validity training direction & Mapped satisfiability direction & Weight & Intended flip & Other-class damage & Unparseable & Interpretation \\
\midrule
Valid \(\rightarrow\) invalid & Sat. \(\rightarrow\) unsat. & 1.0 & 1.4\% & 8.7\% & N/A & Primary \\
Valid \(\rightarrow\) invalid & Sat. \(\rightarrow\) unsat. & 2.0 & 1.8\% & 1.7\% & N/A & Sensitivity \\
Invalid \(\rightarrow\) valid & Unsat. \(\rightarrow\) sat. & 1.0 & 0.3\% & 42.1\% & N/A & Primary \\
Invalid \(\rightarrow\) valid & Unsat. \(\rightarrow\) sat. & 2.0 & 0.0\% & 47.9\% & N/A & Sensitivity \\
\bottomrule
\end{tabular}
}
\caption{Gemma 1/0 transfer from validity to satisfiability. Weight 1.0 matches the main objective and is primary; weight 2.0 is reported separately. Intended flip and destination-task damage are distinct: some prefixes disturb the destination task without implementing the intended answer operation.}
\label{tab:appendix-cross-task}
\end{table}

For Qwen3.6, the strongest aggregate target-aligned transfer is 12.1\% under A/B and 11.8\% under 1/0, still much smaller than the effects within one task. The Qwen and Gemma suites use different interfaces and should not be read as a direct model ranking.

\section{Activation Patching Details}

The unprefixed run uses the ordinary prompt, with no dummy prefix. Both runs are passed through the model using input embeddings and an attention mask; position IDs are not supplied separately. The models therefore assign positions in sequence order, so a learned prefix of length \(m\) shifts every original prompt token forward by \(m\) positions. Prompt token \(i\) in the unprefixed run is aligned by content with position \(i+m\) in the prefixed run.

A forward hook replaces the residual-stream output after one transformer block at the selected aligned positions. The model-specific block paths differ, but the hook is applied at the same point after the chosen block in all three models. Prefix positions are never patched. The answer position is the final context token immediately before the allowed answer continuation.

The three conditions copy the syllogism tokens, the answer token, or every non-prefix prompt token. The last condition includes the instruction, syllogism delimiters, label mapping, and answer cue. It is intentionally broad: it tests whether restoring the prompt context at one layer can interrupt the answer change, not which token or component is responsible. These prompt regions were not patched separately. Per-case rows and all tested layers will be included in the artifact.

\begin{table}[H]
\centering
\small
\setlength{\tabcolsep}{5pt}
\begin{tabular}{@{}lccccc@{}}
\toprule
Model & Cases & Layer & All prompt tokens \(R\) & Answer token \(R\) & Syllogism tokens \(R\) \\
\midrule
Qwen3.6 MoE & 36 & 35\,/\,40 & 0.80 & 0.05 & 0.00 \\
Qwen3-8B & 36 & 32\,/\,36 & 0.82 & 0.43 & 0.00 \\
Gemma 4 31B & 48 & 53\,/\,60 & 0.74 & 0.01 & 0.00 \\
\bottomrule
\end{tabular}
\caption{Restoration at the latest tested nonfinal layer. A value near 1 means that copying the unprefixed activation into the prefixed run removes most of the target-score change. Copying all prompt tokens restores substantially more than copying only the syllogism or answer token.}
\label{tab:appendix-patching-details}
\end{table}

For the Gemma validity high-flip group alone, the 24 cases give restoration of 0.786 for all prompt tokens, 0.012 for the answer token, and approximately zero for the syllogism tokens. The validity all-example-target group gives 0.919 restoration for all prompt tokens and approximately zero for either narrow patch over 12 cases. Shuffled validity prefixes produced no qualifying unprefixed-correct flips, so there were no shuffled validity cases to patch.
\begin{landscape}
\section{Full Activation-Change Results}

Table~\ref{tab:bucket} gives the canonical final-layer values summarized by the main activation figure.

\begin{table}[H]
\centering
\footnotesize
\setlength{\tabcolsep}{4pt}
\resizebox{\linewidth}{!}{%
\begin{tabular}{@{}llrcc@{}}
\toprule
Model & Prefix group & Directions & Answer-token distance & Target-minus-source margin shift \\
\midrule
Qwen3.6 MoE & High-flip learned & 3 & 0.378 & 13.58 \\
Qwen3.6 MoE & All-example target & 4 & 0.299 & 16.41 \\
Qwen3.6 MoE & Shuffled & 3 & 0.138 & 4.04 \\
Qwen3.6 MoE & Random & 3 & 0.081 & $-0.10$ \\
\midrule
Qwen3-8B & High-flip learned & 3 & 0.395 & 19.43 \\
Qwen3-8B & All-example target & 4 & 0.571 & 22.40 \\
Qwen3-8B & Shuffled & 3 & 0.308 & 16.86 \\
Qwen3-8B & Random & 3 & 0.037 & 0.81 \\
\midrule
Gemma 4 31B & High-flip learned & 2 & 0.046 & 18.28 \\
Gemma 4 31B & All-example target & 2 & 0.050 & 20.94 \\
Gemma 4 31B & Shuffled & 2 & 0.017 & 1.34 \\
Gemma 4 31B & Random & 2 & 0.009 & 0.21 \\
\bottomrule
\end{tabular}%
}
\caption{Canonical final-layer activation changes for the four prefix groups. Distances compare prefixed and unprefixed answer-token states. Margin shift is the change in the target-answer score minus the source-answer score. Values aggregate the reported direction-specific activation summaries; they are descriptive and are not calibrated across model families.}
\label{tab:bucket}
\end{table}
\end{landscape}

\begin{figure}[H]
\centering
\includegraphics[width=0.82\linewidth]{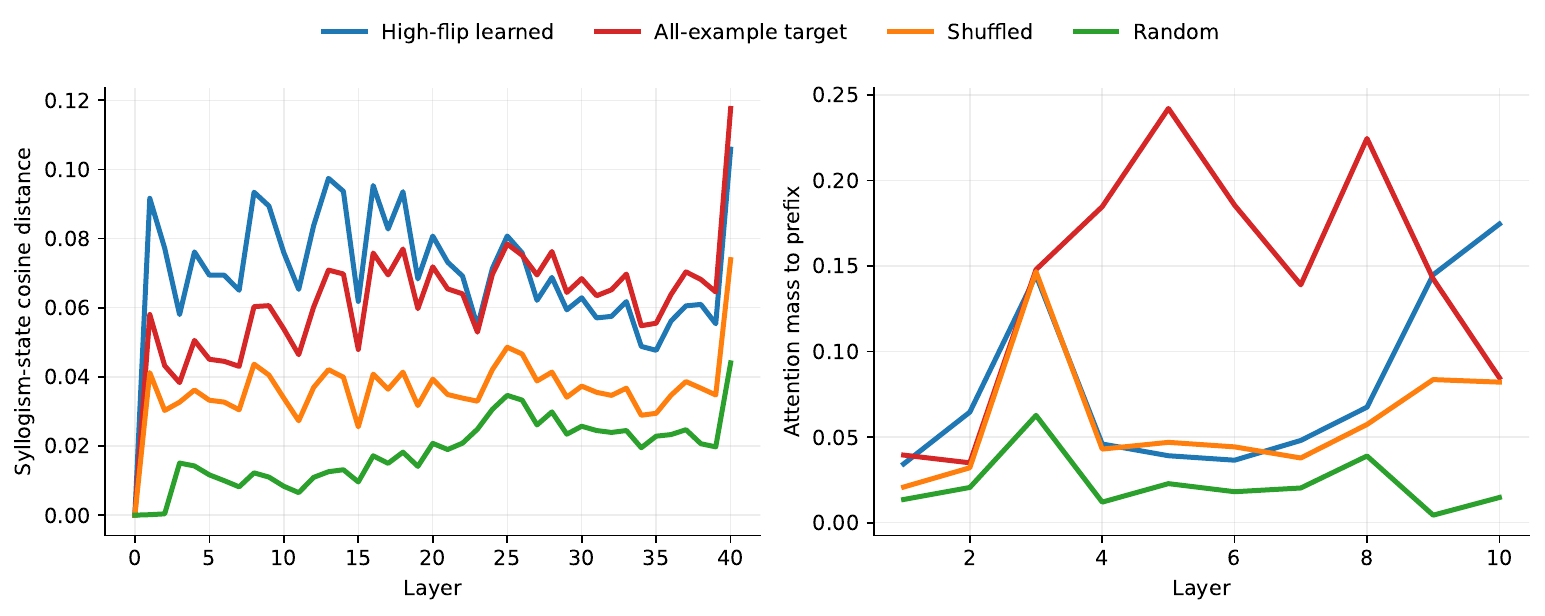}
\caption{Additional layer-wise diagnostics for Qwen3.6 MoE: syllogism-token state distance and attention to the prefix. Answer-token distance and margin shift appear in Figure~\ref{fig:bucket-three-model-mechanism}.}
\label{fig:appendix-qwen36-bucket}
\end{figure}

\begin{figure}[H]
\centering
\includegraphics[width=0.82\linewidth]{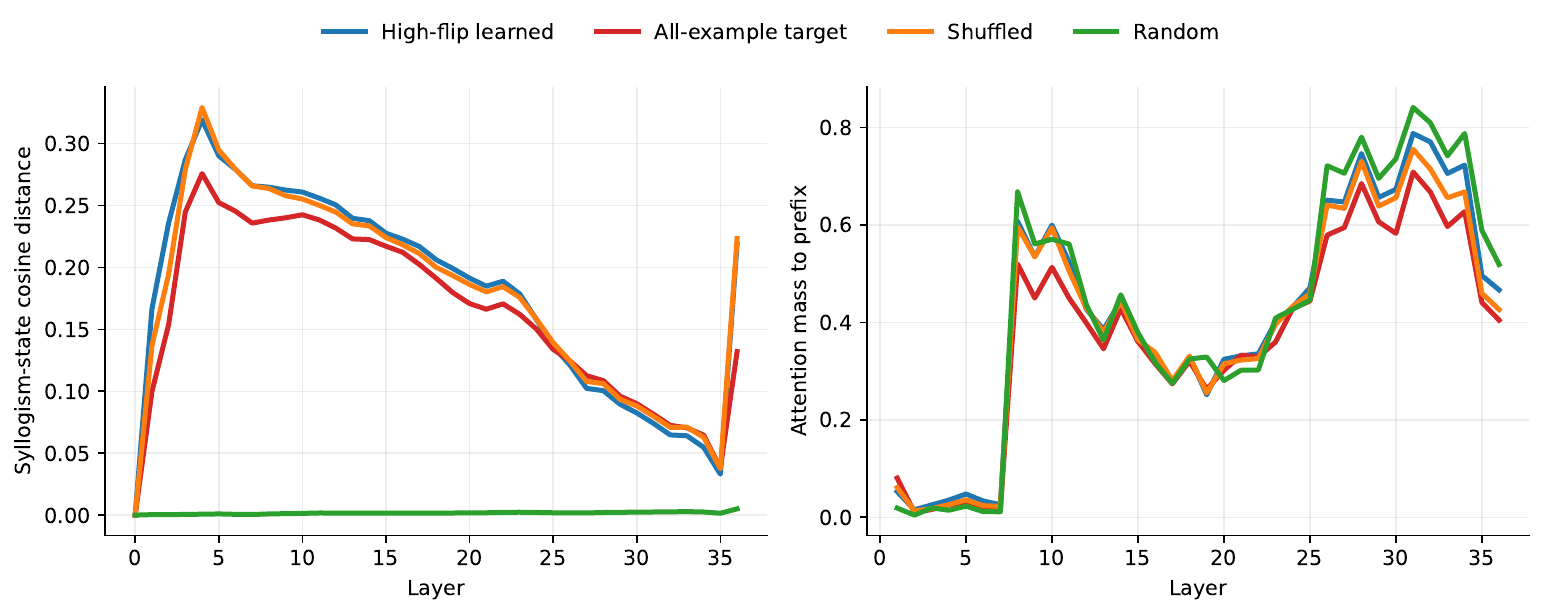}
\caption{Additional layer-wise diagnostics for Qwen3-8B: syllogism-token state distance and attention to the prefix. Answer-token distance and margin shift appear in Figure~\ref{fig:bucket-three-model-mechanism}.}
\label{fig:appendix-qwen8-bucket}
\end{figure}

\clearpage
\begin{figure}[t]
\centering
\includegraphics[width=\linewidth]{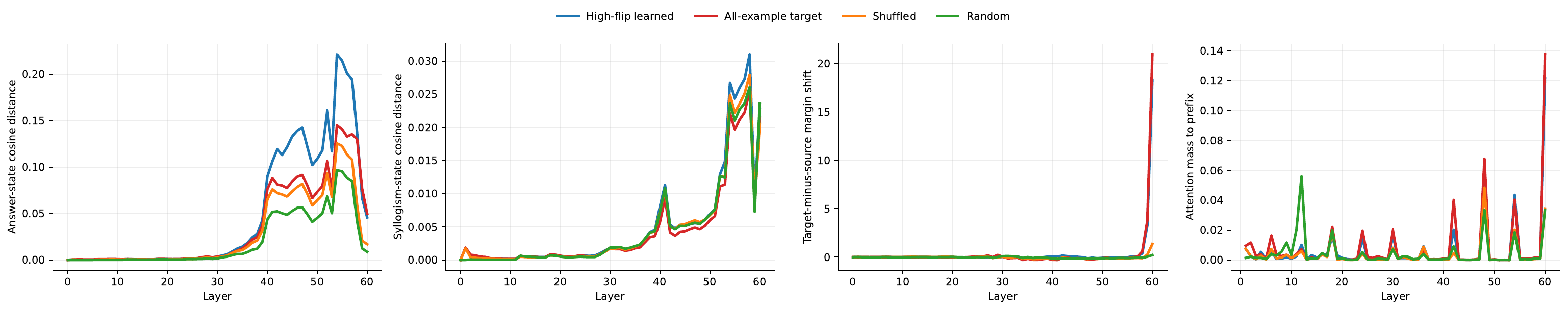}
\caption{Full layer-wise activation diagnostics for Gemma 4 31B: answer-token and syllogism-token state distances, target-minus-source margin shift, and attention to the prefix.}
\label{fig:appendix-gemma-bucket}
\end{figure}

\clearpage
\section{Compute Cost}

Table~\ref{tab:appendix-compute} summarizes the scale of the reported experiments. These are wall-clock ranges from run logs, not normalized hardware benchmarks.

\begin{table}[H]
\centering
\footnotesize
\setlength{\tabcolsep}{4pt}
\begin{tabular}{@{}P{0.24\linewidth}P{0.18\linewidth}P{0.20\linewidth}P{0.28\linewidth}@{}}
\toprule
Stage & GPU use & Wall-clock scale & Notes \\
\midrule
Prompt calibration & up to 8 GPUs & minutes to under 1 hour & Selects a reliable prompt and answer format for each model. \\
\addlinespace
Main prefix experiments & up to 8 GPUs & roughly 0.5--8 hours per broad run & Each prefix uses 100 optimization steps; runtime depends on seeds, wordings, and directions. \\
\addlinespace
Selected-prefix checks & up to 8 GPUs & roughly 30 minutes & Includes intervals, ablations, and generated-answer checks. \\
\addlinespace
Score-model analyses & CPU only & minutes & Fits one-shift, two-shift, affine, margin-bin, and isotonic models for 384 trained prefixes from saved scores; no model inference. \\
\addlinespace
Best-of-1000 random search & up to 8 GPUs & several hours per condition & Random prefixes are matched in norm, ranked on development data, and tested on unseen forms. \\
\addlinespace
Activation analyses & up to 8 GPUs & minutes to under 1 hour per model & Uses selected prefixes and sampled examples rather than retraining. \\
\bottomrule
\end{tabular}
\caption{Approximate compute cost. Exact commands, timestamps, logs, and configuration files will be included in the archival artifact.}
\label{tab:appendix-compute}
\end{table}

\section{Reproducibility Artifacts}

The code, generated data, saved per-example scores, trained prefixes, and analysis outputs will be released in a public archival artifact after the review process. The publication registry records the canonical Gemma validity runs and excludes obsolete prompt artifacts. Each prefix is identified by checkpoint revision, configuration, run, and the SHA-256 hash of \texttt{prefix.pt}, yielding 384 distinct score-analysis prefixes. Table~\ref{tab:appendix-artifacts} lists the planned public contents; exact run identifiers are retained in the experiment log.

\begin{table}[htbp]
\centering
\footnotesize
\setlength{\tabcolsep}{3pt}
\begin{tabular}{@{}P{0.20\linewidth}P{0.35\linewidth}P{0.36\linewidth}@{}}
\toprule
Artifact & Planned public path & Purpose \\
\midrule
Experiment log & \path{documentation/experiment_log.md} & Commands, configurations, completion checks, and result summaries. \\
\addlinespace
Environment & \path{environment/} & Python, PyTorch, Transformers, CUDA, and model-loading versions. \\
\addlinespace
Dataset and splits & \path{data/syllogisms/} & Form enumeration, rendered prompts, verified labels, and train/development/test splits. \\
\addlinespace
Prefix training code & \path{src/prefix_interventions/} & Objectives, scoring, prefix reconstruction, and evaluation. \\
\addlinespace
Run scripts & \path{scripts/run_main_experiments.sh} & Command wrappers for the reported model and control experiments. \\
\addlinespace
Model results & \path{results/qwen36/}, \path{results/qwen3_8b/}, \path{results/gemma4_31b/} & Per-example scores, prefix files, intervals, generated answers, and summaries. \\
\addlinespace
Score models & \path{results/score_models/} & One-shift, two-shift, affine, margin-bin, and isotonic fits; residuals, flip agreement, and bootstrap intervals. \\
\addlinespace
Canonical result registry & \path{results/publication_registry/} & Publication values, source paths, file hashes, exclusions, and verified physical-prefix counts. \\
\addlinespace
Activation analyses & \path{results/activation_profiles/}, \path{results/activation_patching/} & Layer-wise comparisons, patching cases, and restoration summaries. \\
\addlinespace
Neutral text control & \path{results/neutral_text_controls/} & Phrase search, selected development examples, and test results. \\
\bottomrule
\end{tabular}
\caption{Planned contents of the archival artifact. The paths are public-facing names, not machine-local paths.}
\label{tab:appendix-artifacts}
\end{table}

\end{document}